\documentclass[lettersize, journal]{IEEEtran}

\usepackage{amsmath,amssymb,amsfonts}
\usepackage{algorithmic}
\usepackage{graphicx}
\usepackage{tabularx}
\usepackage{textcomp}
\usepackage{wrapfig}
\usepackage{multirow}
\usepackage{subfig}
\usepackage{comment}
\usepackage{booktabs}
\usepackage{soul,xcolor}
\usepackage{ulem}
\usepackage{url}
\usepackage{makecell}
\usepackage[inkscapelatex=false]{svg}
\usepackage{tabularray}
\usepackage[hidelinks]{hyperref}

\newcommand{\sensordata}{\mathbf{x}}
\newcommand{\randvec}{\mathbf{z}}

\usepackage{array}
\newcolumntype{P}[1]{>{\centering\arraybackslash}p{#1}}

\ifCLASSINFOpdf
\else
\fi
\begin{document}
\setstcolor{red}
%
\title{Data-driven Camera and Lidar Simulation Models for Autonomous Driving: A Review from Generative Models to Volume Renderers}
%
%
%

\author{Hamed~Haghighi$^1$,~Xiaomeng~Wang$^1$,~Hao~Jing$^1$,
        and~Mehrdad~Dianati$^2$~
\thanks{$^{1}$H. Haghighi, X. Wang, and H. Jing are with WMG, University of Warwick, Coventry, U.K. (Corresponding author: Hamed.Haghighi@warwick.ac.uk)
}
\thanks{$^{2}$M. Dianati is with the School of Electronics, Electrical Engineering and Computer Science at Queen’s University of Belfast and the WMG at the University of Warwick.
}
}

%
%

\markboth{}%
{Shell \MakeLowercase{\textit{et al.}}: Bare Demo of IEEEtran.cls for IEEE Journals}
%



\maketitle
\begin{abstract}
Perception sensors, particularly camera and Lidar, are key elements of Autonomous Driving Systems (ADS) that enable them to comprehend their surroundings for informed driving and control decisions. Therefore, developing realistic simulation models for these sensors is essential for conducting effective simulation-based testing of ADS. Moreover, the rise of deep learning-based perception models has increased the utility of sensor simulation models for synthesising diverse training datasets. The traditional sensor simulation models rely on computationally expensive physics-based algorithms, specifically in complex systems such as ADS. Hence, the current potential resides in data-driven approaches, fuelled by the exceptional performance of deep generative models in capturing high-dimensional data distribution  and volume renderers in accurately representing scenes. This paper reviews the current state-of-the-art data-driven camera and Lidar simulation models and their evaluation methods. It explores a spectrum of models from the novel perspective of generative models and volume renderers. Generative models are discussed in terms of their input-output types, while volume renderers are categorised based on their input encoding. Finally, the paper illustrates commonly used evaluation techniques for assessing sensor simulation models and highlights the existing research gaps in the area.
\end{abstract}


%
\IEEEpeerreviewmaketitle

\section{Introduction}
\label{sec:introduction}
Safety is crucial in Autonomous Driving Systems (ADS), given the potentially severe consequences of system failures, as evidenced by recent Uber and Tesla crashes \cite{Uber, Tesla}, which highlight the need for stringent testing protocols. Physical testing of ADS, though valuable, is challenging due to the time, labour, and cost involved, and requires extensive miles to statistically validate safety \cite{Kalra2016}. Moreover, certain dangerous scenarios may not be feasible or ethical for real-world testing. Conversely, virtual testing in simulation environments offers several advantages, including the efficient simulation of extensive miles in a short period, testing safety-critical scenarios without physical damage, and modelling complex and costly-to-recreate traffic scenarios.
\begin{figure*}[!t]
\centerline{\includegraphics[width=\textwidth]{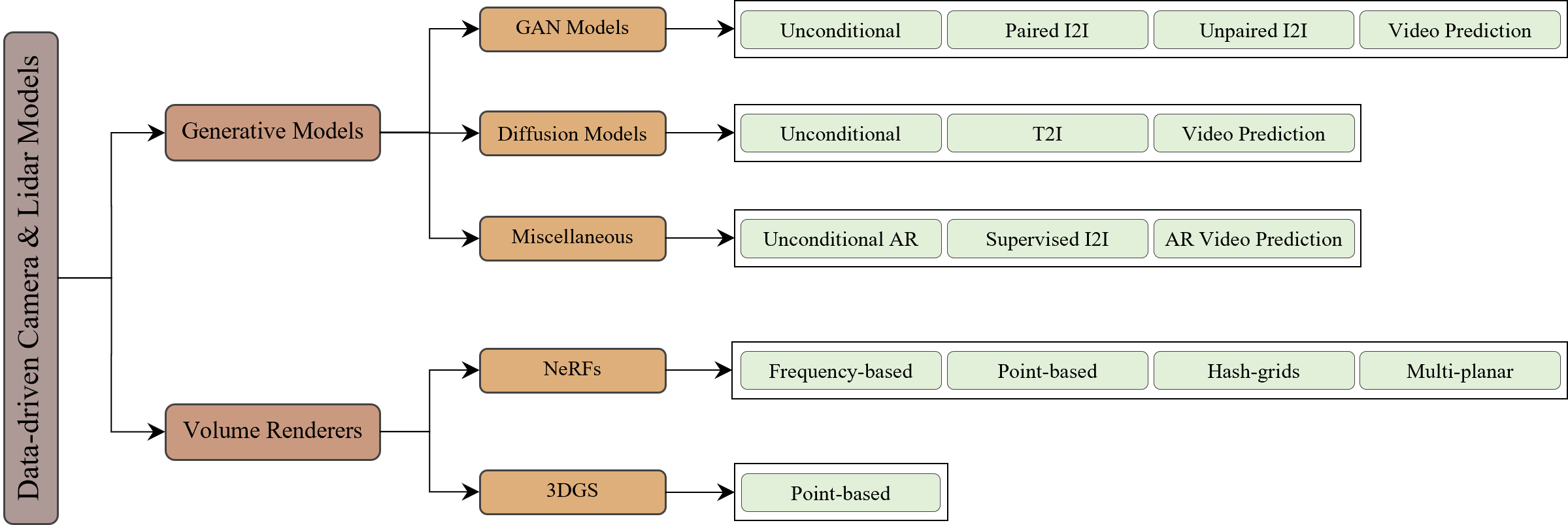}}
\caption{Categorisation of data-driven camera and Lidar simulation models for ADS.}
\label{fig:categorisation}
\end{figure*}
In the domain of ADS applications, perception sensors, including cameras and Lidar, play a pivotal role. These sensors monitor the surrounding environment, detecting moving objects such as vehicles, cyclists, pedestrians, and stationary objects such as traffic lights and road signs. It is critical to test ADS with the realistic performance of the perception sensors, especially in challenging environments where their performances are likely to be degraded. Hence, development of realistic sensor simulation models becomes vital, facilitating extensive testing and validation in simulated environments and contributing to the overall safety and reliability of ADS. \par
The surge in deep learning-based models for ADS, especially in perception applications \cite{Grigorescu2019ASO}, has led to a substantial demand for annotated sensory datasets to train these models. For this reason, numerous sensory datasets have been recorded from real driving scenes \cite{9622929} and have been annotated by human labour in recent years. However, the process of collecting and annotating real-world datasets is costly and presents challenges such as privacy concerns and safety hazards. As a solution, many researchers have turned to simulation environments to generate synthetic datasets. Simulation frameworks enable the rapid creation of extensive sensory data with ground-truth annotations and the generation of edge-case scenarios without posing safety hazards. These synthetic datasets are either used in conjunction with real datasets to train perception models or employed independently with domain adaptation techniques. In both scenarios, realistic simulation of perception sensors plays a significant role in enhancing the performance of downstream perception tasks.

\par
The literature on sensor simulation models presents two fundamental approaches: physics-based and data-driven techniques \cite{sota_sensor}. Physics-based methods involve explicit simulations of sensor-related physical phenomena, relying on intricate hand-crafted formulations for approximations. For instance, Liu et al. \cite{Liu2019} introduced a high-fidelity physics-based camera model for autonomous driving, incorporating components that precisely simulate light propagation, surface materials, camera lens, and aperture. Although capable of generating high-fidelity sensor images, such systems require extensive computations. In contrast, data-driven models have gained popularity in recent years to address the complexity of physics-based models. Unlike physics-based approaches, data-driven models leverage statistical models to implicitly uncover underlying relations by learning from data. The growing availability of real-world recorded perception sensory datasets, coupled with the success of deep generative models in synthesising high-dimensional sensory data and accuracy of volume renderers in 3D scene representation, has led to a rapid expansion of the literature on data-driven models. \par
In this article, we review the State-Of-The-Art (SOTA) data-driven camera and Lidar simulation models, and their evaluation techniques. We propose a novel perspective exploring methods from the standpoint of generative models and volume renderers. Generative models focus on modelling sensory data through implicit distribution approximators such as Generative Adversarial Networks (GANs) \cite{NIPS2014_5ca3e9b1}, denoising diffusion Models~\cite{10.5555/3495724.3496298}, or Auto-Regressive (AR) models. On the other hand, volume renderers adopt a more explicit approach, utilising learning-based models, such as Neural Radiance Fields (NeRFs) \cite{10.1145/3503250} or 3D Gaussian Splatting (3DGS) \cite{10.1145/3592433} models, for scene representation and ray-marching. We further categorise the generative models based on their input-output data into unconditional, Image-to-Image (I2I) translation, Text-to-Image (T2I) translation and video prediction models. Additionally, volume renderers are further categorised based on their input encoding methods into frequency-based, hash-grids, point-based and multi-planar approaches. The categorisation of data-driven camera and Lidar simulation models is shown in Fig. \ref{fig:categorisation}. Furthermore, we review the existing evaluation methods for data-driven sensor simulation models, exploring both qualitative and quantitative approaches.

\par

In the context of ADS simulation, several literature review papers have investigated various aspects, including driving simulation frameworks, synthetic datasets, and sensor simulation approaches. For instance, Rosique et al. \cite{Rosique2019} conducted a review of perception systems and simulators for ADS, emphasising the characteristics of sensor hardware and simulators used in vehicle tests, game engines, and robotics. Kang et al. \cite{kang2019} provided an overview of public driving datasets and virtual testing environments, focusing on accessible virtual testing environments for closed-loop ADS testing. Schlager et al. \cite{sota_sensor} conducted a study reviewing perception sensor models, categorising radar, Lidar, and camera models based on fidelity levels. In a recent work \cite{9564034}, the authors concentrated on reviewing digital camera components and their simulation approaches in the context of ADS and robotics. Despite the coverage of various sensor simulation methods, there is a noticeable gap in the discussion of SOTA data-driven methods, particularly those based on generative models and volume renderers. Furthermore, existing research has not adequately investigated sensor simulation evaluation approaches, a crucial aspect for virtual verification and validation of ADS. The summary of our paper's contributions is as follows:
\begin{itemize}
    \item A comprehensive literature review of data-driven camera and Lidar simulation models is carried out, with a specific emphasis on the latest techniques. 
    \item A novel perspective on data-driven sensor simulation models is discussed, exploring both implicit approaches such as generative models and more explicit models such as volume renderers. 
    \item A detailed explanation and categorisation of evaluation approaches for sensor simulation models are provided. 
\end{itemize}
The structure of this paper is as follows: Section \ref{sec:background} provides background information on data-driven simulation models, emphasising both generative models and volume renderers. Sections \ref{sec:GANs}, \ref{sec:DMs}, and \ref{sec:Misc} delve into various generative models, including GAN-based models, diffusion models and miscellaneous, respectively. Sections \ref{sec:NeRFs} and \ref{sec:3DGS} address volume renderers, focusing on NeRFs and 3DGS models, respectively. Section \ref{sec:eval} outlines methods for the evaluation of simulation models. Finally, Section \ref{sec:conclusion} provides concluding remarks, identifies research gaps, and suggests directions for future research.

\begin{figure*}[!t]
\centerline{\includegraphics[width=\textwidth]{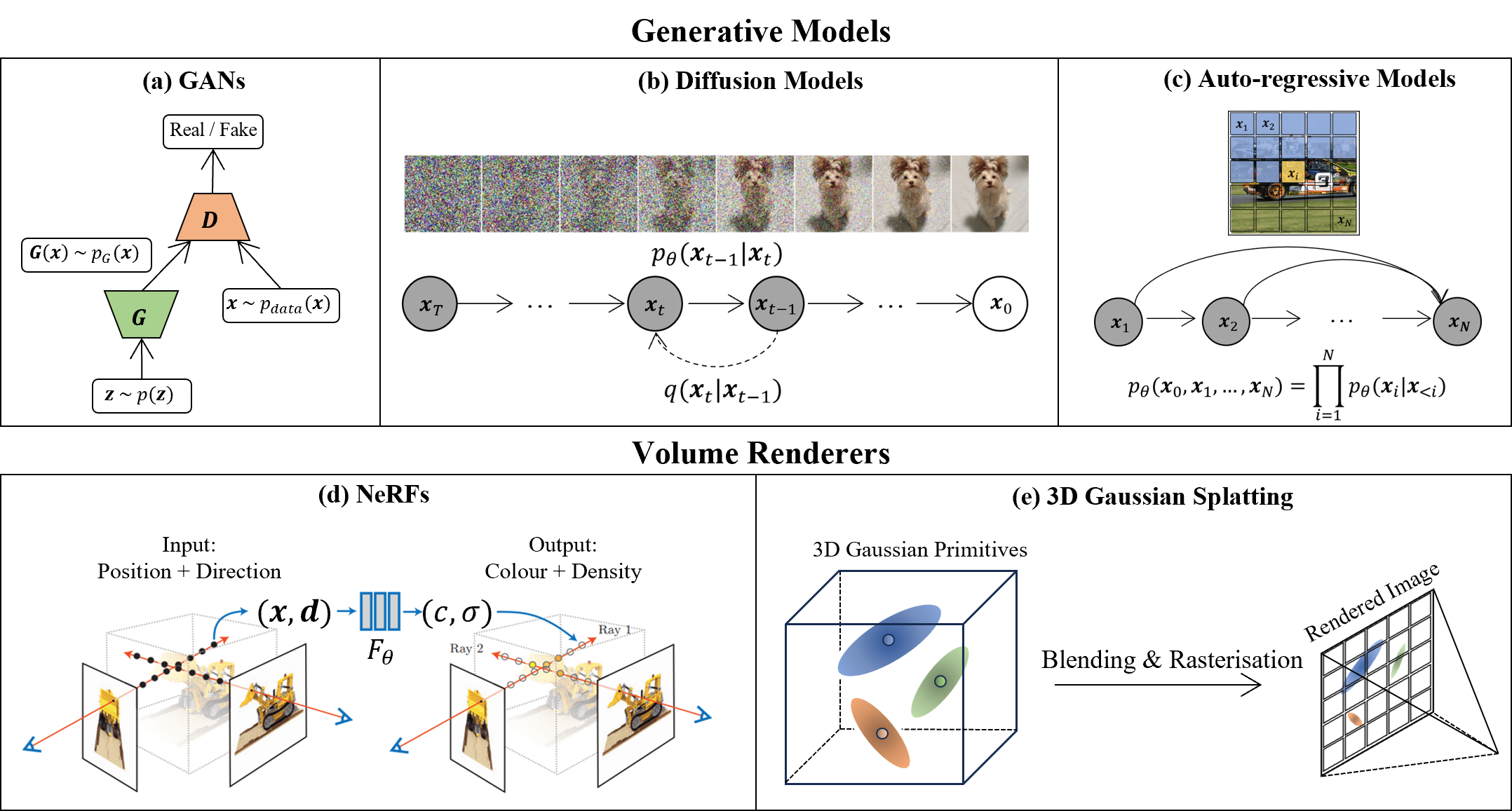}}
\caption{An overview of data-driven models, including generative models and volume renderers, that are widely used for camera and Lidar simulation in ADS. These data-driven models contain generative approaches such as (a) GANs \cite{NIPS2014_5ca3e9b1}, (b) denoising diffusion models \cite{10.5555/3495724.3496298}, and (c) auto-regressive models, while volume renderers include (d) NeRFs \cite{10.1145/3503250} and (e) 3D Gaussian splatting models \cite{10.1145/3592433}. The car image in (c) is sourced from ImageNet \cite{deng2009imagenet}.}
\label{fig:background}
\end{figure*}
\section{Background} \label{sec:background}

Data-driven approaches model sensory data by leveraging large real-world datasets to train models that predict or simulate new sensory outputs with high accuracy. Unlike traditional methods that rely on manual formulas and exact physics-based simulations, data-driven methods abstract patterns directly from data, enabling them to dynamically adapt to complex and variable real-world conditions. This capability significantly enhances the scalability and realism of simulations.

In the landscape of data-driven modelling, generative models and volume rendering models represent two distinct paradigms. Generative models, including Generative Adversarial Networks (GANs) \cite{NIPS2014_5ca3e9b1} and denoising diffusion models~\cite{10.5555/3495724.3496298}, learn statistical data distributions to implicitly generate new, plausible data instances. They excel in producing diverse scenarios without explicitly defining every possible condition, making them ideal for generating diverse sensory data. In contrast, volume renderers such as Neural Radiance Fields (NeRFs) \cite{10.1145/3503250} provide a representation of the environment. They construct highly detailed and accurate 3D reconstructions by mapping spatial coordinates to visual properties, which can be rendered from any viewpoint. This distinction paves the way for a detailed exploration of each method's specific techniques and advantages, focusing on how they address different needs in simulation.

\subsection{Generative Models}
Generative models attempt to learn the distribution of real-world data and generate new instances that mimic this distribution without the need for explicit scene geometry or physical properties. In the context of sensor simulation, generative models aim to approximate a target probability distribution \( p_{\text{data}}(\sensordata) \) derived from a set of observed sensory data \( \mathbf{\sensordata} \in \mathbb{R}^N \), where \( N \) is the dimensionality of the sensor output, The goal is to learn a parameterised model distribution \( p_{\theta}(\mathbf{\sensordata}) \) that closely represents \( p_{\text{data}}(\mathbf{\sensordata}) \). Training a generative model involves adjusting the parameters \( \theta \) of the model such that \( p_{\theta}(\sensordata) \) becomes an effective approximation of \( p_{\text{data}}(\sensordata) \). Generative models may employ a variety of strategies depending on their architecture and the specific training regime. Techniques such as adversarial training in GANs or the iterative refinement in diffusion models are used to align \( p_{\theta}(\sensordata) \) with \( p_{\text{data}}(\sensordata) \), emphasising model flexibility in capturing and reproducing the complex statistical properties of sensory data.

\subsubsection{Generative Adversarial Networks (GANs)}

Generative Adversarial Networks (GANs) \cite{NIPS2014_5ca3e9b1} is a form of generative model, which consists of a generator (\(G\)) and a discriminator (\(D\)). The generator \(G\) attempts to generate data that mimics real-world data, learning to map from a latent space $z \in \mathbb{R}^M (M < N)$  to the data space, aiming to match the real data distribution \(p_{\text{data}}(\sensordata)\). In contrast, the discriminator \(D\) evaluates the realism of samples, tasked with distinguishing real data from that generated by \(G\), as shown in Fig. \ref{fig:background} (a).

The interaction between \(G\) and \(D\) is formulated as a mini-max game, represented by the following equation:
\begin{equation}
\begin{split}
       \min_{G} \max_{D} \mathcal{L}(D, G) &= \mathbb{E}_{\sensordata \sim p_{\text{data}}(\sensordata)}\log D(\sensordata)\\ &+ \mathbb{E}_{z \sim p(\randvec)}\log(1 - D(G(\randvec))).
\end{split}
\end{equation}
Advantages of GANs include their ability to generate high-quality, realistic images, which makes them particularly useful for applications requiring high visual fidelity, such as automotive simulation. However, challenges with GANs include their training instability. The adversarial nature of their training can lead to issues such as non-convergence and mode collapse, where \(G\) fails to produce diverse outputs and instead generates repetitive or overly similar samples. Moreover, the sensitivity of GANs to the settings of hyper-parameters requires careful tuning to achieve optimal performance and stability.

\subsubsection{Diffusion Models}
Denoising diffusion probabilistic models \cite{10.5555/3495724.3496298}, referred to as Diffusion Models (DMs), represent another class of generative models that operate by progressively adding and then removing noise to generate data. These models simulate the forward process where noise is incrementally added to the data \( \sensordata_{0} \) from the real data distribution \( p_{data}(\sensordata_{0}) \), resulting in a gradually noisier dataset over a sequence of steps \( \sensordata_{1},\sensordata_{2}, ..., \sensordata_{T} \). The reverse process involves a model that learns to denoise this data, effectively reconstructing the data back towards its original form (See Fig. \ref{fig:background} (b)).
\begin{table*}
\centering
\caption{Comparison of widely used data-driven simulation methods in ADS.}
\label{tab:comparison}
\begin{tblr}{
colspec = {Q[100,l]Q[110,c]Q[80,c]Q[60,c]Q[60,c]Q[80,c]Q[80,c]Q[80,c]},
row{1} = {font=\bfseries},
}
\hline
Method & Category & 3D Scene \newline Representation & Output Realism & Inference Speed & Long-range Modelling & Training Stability & Output Diversity \\
\hline
\SetCell[r=3]{l} Generative Models
& GANs \cite{NIPS2014_5ca3e9b1} & $\times$ & **\textcolor{lightgray}{*} & *** & *\textcolor{lightgray}{**} & *\textcolor{lightgray}{**} & **\textcolor{lightgray}{*} \\
\cline{2-8}
& Diffusion Models \cite{10.5555/3495724.3496298} & $\times$ & *** & **\textcolor{lightgray}{*} & **\textcolor{lightgray}{*} & *** & *** \\
\cline{2-8}
& AR Models & $\times$ & **\textcolor{lightgray}{*} & **\textcolor{lightgray}{*} & *** & **\textcolor{lightgray}{*} & *** \\
\hline
\SetCell[r=2]{l} Volume Renderers
& NeRFs \cite{10.1145/3503250} & $\checkmark$ & *** & **\textcolor{lightgray}{*} & **\textcolor{lightgray}{*} & **\textcolor{lightgray}{*} & *\textcolor{lightgray}{**} \\
\cline{2-8}
& 3DGS \cite{10.1145/3592433} & $\checkmark$ & **\textcolor{lightgray}{*} & *** & **\textcolor{lightgray}{*} & **\textcolor{lightgray}{*} & *\textcolor{lightgray}{**} \\
\hline
\end{tblr}
\end{table*}
The mathematical formulation of DMs is centred around the concept of reversing the noise addition process. The generative process models the conditional probability of recovering a previous state \( x_{t-1} \) from a noisier state \( x_t \), described by:
\begin{equation}
    p_{\theta}(\sensordata_{t-1} | \sensordata_t) = \mathcal{N}(\sensordata_{t-1}; \mu_{\theta}(\sensordata_{t}, t), \Sigma_{\theta}(\sensordata_{t}, t)),
\end{equation}
where \( \mu_{\theta} \) and \( \Sigma_{\theta} \) are functions parameterised by the model that estimate the mean and variance needed to denoise \( x_t \).
\par
The optimisation of DMs is typically approached by minimising a variational lower bound, often resulting in a simplified objective such as the mean squared error between the denoised and original data across the diffusion steps. The loss function generally used can be expressed as:
\begin{equation}
    \mathcal{L}(\theta) = \mathbb{E}_{t, \sensordata_{0}, \epsilon} \left[ ||\epsilon - \epsilon_{\theta}(\sensordata_{t}, t)||^2 \right],
\end{equation}
where \( \epsilon \) is the noise added at each step, and \( \epsilon_{\theta} \) is the noise model predicted by the network.

Diffusion models are known for their ability to generate high-quality and diverse outputs. Their stability in training is a significant advantage over other generative models, as they are less prone to issues like mode collapse, where the generator fails to capture the diversity of the dataset. However, a notable challenge with diffusion models is their computational efficiency. The iterative nature of the reverse process, which requires multiple network evaluations to generate a single sample, makes them computationally intensive compared to models that generate outputs in a single forward pass.
\subsubsection{Auto-regressive Models}
Auto-regressive (AR) models form another important category of generative models used for data-driven sensor simulation. These models generate data dimensions sequentially, where each data dimension is conditioned on the previous ones.
In an AR model, the probability of a data \( \sensordata = (\sensordata_1, \sensordata_2, ..., \sensordata_N) \) is factorised into a product of conditional probabilities:
\begin{equation}
p(\sensordata) = \prod_{i=1}^{N} p(\sensordata_i | \sensordata_{<i}),
\end{equation}
where \( \sensordata_{<i} \) represents all the previous dimensions before \( i^{th}\) dimension (See Fig. \ref{fig:background} (c)).

Training an AR model involves maximising the likelihood of the training data under the model. This is typically done by minimising the negative log-likelihood:
\begin{equation}
\mathcal{L}(\theta) = -\sum_{i=1}^{T} \log p_{\theta}(\sensordata_t | \sensordata_{<t}).
\end{equation}
Auto-regressive models can be implemented using architectures such as Recurrent Neural Networks (RNNs), Long Short-Term Memory networks (LSTMs) \cite{hochreiter1997long}, and transformers \cite{10.5555/3295222.3295349}. These architectures effectively capture long-range dependencies and complex patterns in the data. One key advantage of AR models is their ability to generate data with high fidelity and temporal coherence, making them well-suited for tasks such as video prediction and time-series forecasting. However, the sequential nature of AR models can lead to slower inference times, as each data point must be generated step-by-step.

\subsection{Volume Renderers}
Volume rendering is a significant technique in computer graphics for visualising volumetric data, which involves simulating the propagation of light through a three-dimensional space. Contrary to generative models that indirectly encode the complexity of an environment by learning its data distribution, volume rendering models use explicit spatial information. They reconstruct scenes by integrating the interaction of light with the volumetric attributes of the environment, such as density and colour at every point in a three-dimensional space. This explicit approach is especially advantageous for simulations requiring precise optical and physical realism, as it can accurately depict how light interacts within complex environments.

The foundational principle of volume rendering can be captured by the volume rendering equation, which computes the contribution of light absorbed and emitted as it travels through a volume. The general formulation is:

\begin{equation}
    C(\mathbf{r}) = \int_{t_0}^{t_1} T(t) \sigma(\mathbf{x}(t)) \mathbf{c}(\mathbf{x}(t), \mathbf{d}) \, dt,
\end{equation}
where \( C(\mathbf{r}) \in \mathbb{R}\) denotes the colour accumulated along ray \( \mathbf{r} \in \mathbb{R}^3 \), \( \sigma(\mathbf{x}(t)) \in \mathbb{R}  \) is the volume density at point \( \mathbf{x}(t) \in \mathbb{R}^3  \), \( \mathbf{c}(\mathbf{x}(t), \mathbf{d}) \in \mathbb{R}^3 \) represents the emitted colour at that point dependent on direction \( \mathbf{d} \in \mathbb{R}^3  \), and \( T(t) \in \mathbb{R} \) is the transmittance, representing the light's attenuation from the start of the ray at \( t_0 \in \mathbb{R} \) to \( t \in \mathbb{R} \), calculated as \( \exp(-\int_{t_0}^t \sigma(\mathbf{x}(s)) \, ds) \). This equation allows for detailed simulation of lighting effects through different materials, making volume rendering particularly effective.
\subsubsection{Neural Radiance Fields (NeRFs)}
Neural Radiance Fields (NeRFs) \cite{10.1145/3503250} are a type of volume renderers, that utilise continuous scene representations; They optimise a Multi-Layer Perceptron (MLP) to simulate light interactions within 3D environments through volumetric ray-marching. The fundamental NeRF formulation integrates radiance along a camera ray, calculating colour as a function of accumulated light and material properties at each point along the ray:
\begin{equation}
    C(\mathbf{r}) = \int_{t_0}^{t_1} T(t) \sigma_{\theta}(\mathbf{x}(t)) \mathbf{c}_{\theta}(\mathbf{x}(t), \mathbf{d}) \, dt.
\end{equation}
This integration considers both the emitted color \( \mathbf{c}_{\theta} \) and the density \( \sigma_{\theta} \), which are output by a MLP (See Fig. \ref{fig:background} (d)). The transmittance \( T(t) \) describes how much light survives without being scattered or absorbed, crucial for the realistic rendering of materials like fog, smoke, or translucent objects. NeRFs excel at producing photo-realistic images and are capable of synthesizing new views from limited datasets. However, they require extensive computational resources for training and inference and are best suited for static scenes.

\subsubsection{3D Gaussian Splatting}
3D Gaussian Splatting (3DGS) \cite{10.1145/3592433} introduces a novel approach to volume rendering that merges the explicit representation advantages of traditional methods, such as meshes and points, with the continuous scene modelling of NeRFs. By employing a tile-based splatting mechanism with anisotropic 3D Gaussian, this method achieves the flexibility required for real-time rendering (see Fig. \ref{fig:background} (e)).
The 3DGS utilise 3D points characterised by specific features such as colour $\mathbf{c}_{i} \in \mathbb{R}$, weight $\sigma_i \in \mathbb{R}$, mean \(\mathbf{p}_i \in \mathbb{R}^3\), and covariance \(\Sigma_i \in \mathbb{R}^{3 \times 3}\) to represent the scene. It then projects the points on the image plane and transforms the features based on this projection to calculate the final image intensity. The Gaussian weight in the image plane for a point \(\mathbf{x} \in \mathbb{R}^2\) is as:
\begin{equation}
    \sigma_i(\mathbf{x}) = \frac{1}{\left|\mathbf{J}_i^{-1}\right|} \exp\left(-\frac{1}{2} (\mathbf{x} - \mathbf{x}_i)^T (\mathbf{J}_i \Sigma_i \mathbf{J}_i^T)^{-1} (\mathbf{x} - \mathbf{x}_i)\right),
\end{equation}
where \(\mathbf{J}_i \in \mathbb{R}^{2 \times 3}\) is the Jacobian matrix that transforms the 3D point \(\mathbf{p}_i \) during its projection onto the 2D image plane, resulting in \(\mathbf{x}_i\). This transformation adapts the Gaussian shape to accommodate the camera’s perspective effects.

The final image intensity at a pixel position \(\mathbf{x}\) in the 2D plane is derived by integrating the contributions from all points weighted by their respective Gaussian distributions:
\begin{equation}
    c(\mathbf{x}) = \frac{\sum_{i=1}^{N} \sigma_i(\mathbf{x}) \cdot \mathbf{c}_i}{\sum_{i=1}^{N} \sigma_i(\mathbf{x})}.
\end{equation}
This formulation ensures that the spatial and depth cues from the 3D scene are accurately represented in the 2D image, providing a realistic rendering of the scene according to the camera optics and geometry.
\begin{table*}
\centering
\caption{Summary of unconditional GAN-based Models.}
\label{tab:uc_gan}
\begin{tblr}{
colspec = {Q[120,c]Q[50,c]Q[20,c]Q[400,l]Q[120,l]},
}
\hline
\textbf{Model} & \textbf{Sensor} & \textbf{Year} & \textbf{Description} & \textbf{Datasets} \\ \hline
LidarGAN \cite{Caccia2018DeepGM} & Lidar & 2018 & Synthesises high-quality Lidar scans using deep generative models & KITTI \cite{Geiger2013IJRR} \\ \hline
SB-GAN \cite{49041} & Camera & 2019 & Generates semantic label maps first, then synthesises the final image & Cityscapes \cite{Cordts2016Cityscapes} \\ \hline
Volokitin et al. \cite{9151072} & Camera & 2020 & Separates image generation into layout prediction and detailed image synthesis & Cityscapes 
\cite{Cordts2016Cityscapes} \\ \hline
Semantic Palette \cite{9577811} & Camera & 2021 & Uses class proportions to guide the generative process for scene elements & Cityscapes \cite{Cordts2016Cityscapes} \\ \hline
DUSty \cite{Nakashima2021LearningTD} & Lidar & 2021 & Uses a noise-aware GAN framework to handle dropped points in Lidar scans & KITTI \cite{Geiger2013IJRR}, MPO \cite{Nakashima2021LearningTD} \\ \hline
Dusty-2 \cite{nakashima2022generative} & Lidar & 2022 & Focuses on data-level domain transfer for Lidar range images & Raw KITTI \cite{Geiger2013IJRR} \\ \hline
Urban-StyleGAN \cite{eskandar2023urbanstylegan} & Camera & 2023 & Enables detailed manipulation of urban scene images with high fidelity & Cityscapes \cite{Cordts2016Cityscapes} \\ \hline
\end{tblr}
\end{table*}

\subsection{Comparative Analysis of Methods} 
This section comprehensively compares the discussed data-driven simulation methods, highlighting their key characteristics and capabilities. Table \ref{tab:comparison} summarises these attributes using a scoring system ranging from 1 to 3, represented by '*' symbols. A higher score indicates the superior performance of the model in the respective aspect.
\subsubsection{3D Scene Representation}
Volume renderers, specifically NeRFs and 3DGS, inherently represent complex 3D scenes, synthesising novel views and capturing spatial relationships. In contrast, generative models such as GANs, diffusion models, and auto-regressive models typically focus on capturing data distributions rather than accurately representing 3D scenes. While powerful and versatile, these generative models lack the inherent 3D representation capabilities that make volume renderers particularly suited for novel view synthesis tasks.
\subsubsection{Output Realism and Inference Speed}
Diffusion models and NeRFs demonstrate superior performance in terms of output realism, producing highly detailed and accurate results. However, this quality often comes at the cost of computational efficiency. 3DGS offers an attractive compromise, providing slightly less realism but with significantly faster inference speed than NeRFs. GANs, while potentially sacrificing some realism compared to diffusion models, excel in inference speed. This makes them one of the fastest methods for generating outputs, particularly useful in applications where real-time performance is crucial.
\subsubsection{Long-range Modelling}
Auto-regressive models excel in capturing and modelling long-range dependencies in data, allowing them to generate coherent sequences and maintain consistency over extended outputs. Diffusion models and volume renderers demonstrate moderate capabilities in this aspect, balancing local detail with broader context. GANs, however, typically struggle with long-range consistency, potentially limiting their effectiveness in tasks requiring extended coherence.
\subsubsection{Training Stability}
Diffusion models offer the highest level of training stability among the compared methods, demonstrating resilience to hyperparameter and architecture changes. NeRFs and 3DGS also demonstrate good stability during the training process, contributing to their effectiveness in 3D scene representation. GANs, while powerful, are known for their training instabilities and potential issues like mode collapse. This can make the training process more challenging and potentially less predictable. Among generative models, auto-regressive models generally offer moderate training stability.
\subsubsection{Output Diversity}
Diffusion models and auto-regressive models excel in generating diverse outputs, capable of producing a wide range of variations in their results. GANs can also generate diverse outputs, although they may sometimes suffer from mode collapse, potentially reducing diversity. NeRFs and 3DGS, being primarily focused on scene representation, typically offer lower output diversity. This is because they are usually trained on and reproduce specific scenes, prioritising accuracy over variation.
\par
The comparison of these different characteristics highlights the distinct strengths and weaknesses of each method. Diffusion models stand out for their balance of high-quality outputs, diversity, and training stability. Auto-regressive models demonstrate particular strength in long-range modelling and output diversity. GANs offer the advantage of rapid inference, albeit with potential instabilities. NeRFs and 3DGS excel in 3D scene representation and realism, though they may have limitations in their broader generative capabilities. When selecting among these methods, the decision should be guided by the specific requirements and priorities of the task at hand, carefully weighing factors such as output quality, processing speed, diversity of results, and the dimensionality of the required representations.
\begin{figure}[!t]
\centerline{\includegraphics[width=\linewidth]{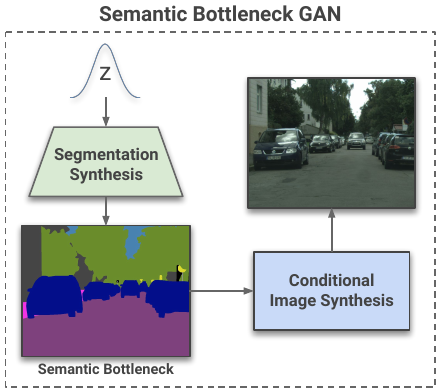}}
\caption{Two-stage synthesis pipeline of the Semantic Bottleneck-GAN \cite{49041}, an unconditional GAN model for RGB image synthesis. }
\label{fig:UC_GAN}
\end{figure}
\section{Generative Models -- GANs} \label{sec:GANs}
GAN-based models constitute the largest category of data-driven approaches for sensor simulation in ADS. The GAN framework relies on the dynamic interaction between two key components: a generator that synthesises data and a discriminator that assesses its realism. Initially designed for unconditional data generation from random noise, GANs have evolved to encompass conditional variants. These variants incorporate additional inputs to guide the data generation process, enabling different applications from both paired and unpaired image-to-image translation to spatio-temporal video prediction. The subsequent sections will explore each category of GAN models for ADS in greater detail.

\subsection{Unconditional Models}
Unconditional GAN models aim to approximate the distribution of observed sensory data without any external guidance or conditions. These models primarily concentrate on capturing the diversity of the data distribution while generating realistic samples. While not typically used directly for sensor simulation in ADS, they serve as foundational frameworks for more complex models. The summary of unconditional GAN models is provided in Table \ref{tab:uc_gan}. Below, we separately dig deep into unconditional models for camera RGB images and Lidar point cloud synthesis.
\begin{table*}[!t]
\centering
\caption{Summary of paired I2I translation GAN-based models (all used for camera RGB image synthesis).}
\label{tab:paired_i2i_models}
\begin{tblr}{
  colspec = {Q[100,c]Q[50,c]Q[470,l]Q[120,l]},
}
\hline
\textbf{Model}  & \textbf{Year} & \textbf{Description} & \textbf{Datasets} \\ \hline
Pix2Pix \cite{Isola2016ImagetoImageTW}  & 2016 & Aligns input semantic label maps with corresponding images using cGAN and L1 losses & Cityscapes \cite{Cordts2016Cityscapes} \\ \hline
Pix2PixHD \cite{Wang2017HighResolutionIS}  & 2017 & Incorporates multi-scale generator and discriminator for high-resolution image synthesis & Cityscapes \cite{Cordts2016Cityscapes} \\ \hline
SPADE \cite{8953676}  & 2019 & Introduces spatially-adaptive normalisation layers for semantic map-based modulation & Cityscapes \cite{Cordts2016Cityscapes} \\ \hline
CC-FPSE \cite{10.5555/3454287.3454339}  & 2019 & Predicts layout-to-image conditional convolution kernels for image generation & Cityscapes \cite{Cordts2016Cityscapes} \\ \hline
SEAN \cite{Zhu_2020_CVPR}  & 2019 & Allows per-region style control using Semantic Region-Adaptive Normalization & Cityscapes \cite{Cordts2016Cityscapes} \\ \hline
SurfelGAN \cite{Yang_2020_CVPR}  & 2020 & Combines texture-mapped surfels and GANs for realistic camera image generation & Waymo \cite{Sun_2020_CVPR} \\ \hline
OASIS \cite{schonfeld2021you}  & 2020 & Redesigns the discriminator to use semantic label maps directly as ground-truth & Cityscapes \cite{Cordts2016Cityscapes} \\ \hline
ECGAN \cite{tang2023edge}  & 2020 & Uses edge information for enhancing semantic consistency and detail in image synthesis & Cityscapes \cite{Cordts2016Cityscapes} \\ \hline
Robusta \cite{hariat2023learning}  & 2023 & Enhances robustness of semantic segmentation models with high-quality perturbed images & Cityscapes \cite{Cordts2016Cityscapes} \\ \hline
\end{tblr}
\end{table*}
\subsubsection{RGB Image synthesis}
Synthesising RGB images from scratch is challenging due to the multi-object and highly diverse nature of driving scenes. To address this, unconditional models often separate the process into generating the semantic layout of the scene first, followed by the RGB image. The following section chronologically reviews papers in this domain, all of which conduct their experiments on the Cityscapes \cite{Cordts2016Cityscapes} dataset.

The Semantic Bottleneck GAN (SB-GAN) \cite{49041} and the decomposed synthesis Model \cite{9151072} introduce novel approaches for generating urban scenes through a two-step process. They employ a two-step method where a semantic label map is first generated unconditionally and then used to guide the synthesis of the final image (as shown in Fig. \ref{fig:UC_GAN}), enhancing the generation of complex, high-resolution images with coherent global structures.

Progressing further, the introduction of the Semantic Palette Model \cite{9577811} marks a significant innovation by using class proportions to guide the generative process. This approach allows users to specify the desired distribution of scene elements, providing unprecedented control over the class composition of generated images. This model greatly enhances the flexibility and practicality of GANs, making them more adaptable to varied application requirements.

The most recent advancement, Urban-StyleGAN \cite{eskandar2023urbanstylegan} enables the generation and detailed manipulation of urban scene images with remarkable fidelity and control. This model incorporates various techniques for managing and manipulating the latent space, providing tools for precise image editing that push the boundaries of realism in generated urban landscapes.

\subsubsection{Lidar Point Cloud Synthesis}
The LidarGAN model \cite{Caccia2018DeepGM} adapts deep generative models such as VAEs \cite{Kingma2014} and GANs to synthesise high-quality Lidar scans by unrolling them into 2D point maps. This method not only generates realistic samples but also learns meaningful latent representations of the data. By augmenting the 2D signal with absolute positional information, the model enhances robustness against noisy and incomplete input data.
\begin{figure*}[!t]
\centerline{\includegraphics[width=\linewidth]{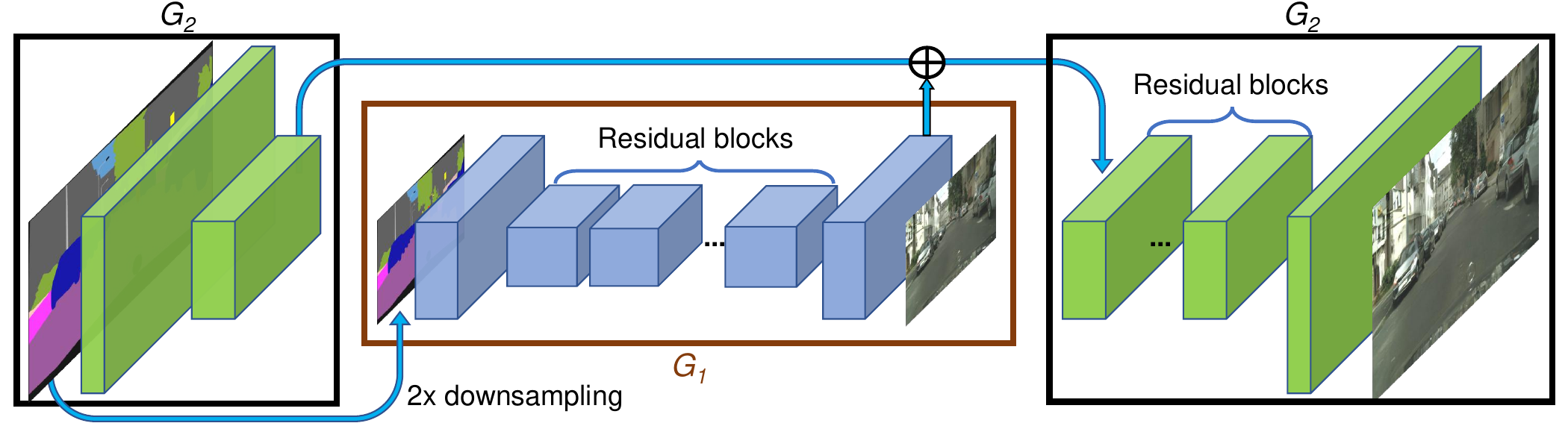}}
\caption{The multi-scale synthesis process of pix2pixHD \cite{Wang2017HighResolutionIS}, a paired I2I model based on GANs. }
\label{fig:I2I_GAN}
\end{figure*}
Building on this foundation, the DUSty model \cite{Nakashima2021LearningTD} introduces a noise-aware GAN framework that tackles the challenge of dropped points in Lidar scans caused by measurement uncertainty. DUSty incorporates a differentiable sampling framework to simulate dropout noises, thereby enhancing the generation of high-quality Lidar data from incomplete observations.

Further advancing, the Dusty-2 model \cite{nakashima2022generative} proposes a generative approach for Lidar range images, focusing on data-level domain transfer and addressing issues such as inconsistent angular resolution and missing properties. This model uses an implicit image representation-based GAN coupled with a differentiable ray-drop effect to enhance the fidelity and diversity of the generated data. Its effectiveness is showcased in tasks such as upsampling, data restoration, and sim-to-real semantic segmentation.

\subsection{Paired Image-to-Image-Translation Models}
Paired Image-to-Image translation (I2I) models are a specific type of conditional model where the input image (condition) and the desired output (ground-truth) are directly paired in the training dataset. In the context of ADS,  these models are typically trained using pairs, such as a semantic segmentation layout with its corresponding RGB image, known as semantic image synthesis. While it is feasible to train paired I2I models with supervised learning, incorporating the GAN framework significantly enhances the realism of the synthesised images. For example, the Pix2Pix model \cite{Isola2016ImagetoImageTW} aligns each input semantic label map with its corresponding image using both the conditional GAN loss \(\mathcal{L}_{\text{cGAN}}\) and the L1 loss \(\mathcal{L}_{\text{L1}}\), defined as:
\begin{align}
    &\mathcal{L}_{\text{Pix2Pix}}(G, D) = \mathcal{L}_{\text{cGAN}}(G, D) + \lambda \mathcal{L}_{\text{L1}}(G)\\
    &\mathcal{L}_{\text{cGAN}}(G, D) = \mathbb{E}_{\sensordata,\mathbf{y}}[\log D(\sensordata, \mathbf{y})] + \mathbb{E}_{\sensordata}[\log(1 - D(\sensordata, G(\sensordata)))] \nonumber\\
    &\mathcal{L}_{\text{L1}}(G) = \mathbb{E}_{\sensordata,\mathbf{y}}[\|\mathbf{y} - G(\sensordata)\|_1] \nonumber,
\end{align}
where \( \sensordata \in X \) represents the input image, \( \mathbf{y} \in Y \) represents the ground-truth image, and \( \lambda \) is a weighting factor that balances two losses. The adversarial loss drives the generator to produce images indistinguishable from real ones, while the L1 loss reduces blurring, promoting sharpness.  The following reviews focus on I2I models for camera RGB image synthesis, all of which have been evaluated on the Cityscapes dataset unless otherwise specified. The summary of paired I2I translation models based on GANs is provided in Table \ref{tab:paired_i2i_models}.

Building on Pix2pix model, Pix2PixHD \cite{Wang2017HighResolutionIS} model addresses the limitations of the original Pix2Pix by incorporating a multi-scale generator and discriminator architecture, as shown in Fig. \ref{fig:I2I_GAN}. This allows for the synthesis of high-resolution images with more detail and texture from semantic maps, thus overcoming the constraints of generating high-quality results from low-resolution inputs.

The innovation continues with the SPADE (Spatially-Adaptive Normalisation) \cite{8953676} model, which introduces spatially-adaptive normalisation layers that modulate activations based on the input's semantic segmentation map. This effectively preserves and utilises semantic information throughout the synthesis process, setting a new standard for photo-realistic and high-resolution image synthesis.
Advancing further, the CC-FPSE \cite{10.5555/3454287.3454339} model introduces the concept of predicting layout-to-image conditional convolution kernels, allowing the semantic layout to directly influence the image generation process. By learning spatially-varying convolution kernels based on the input semantic layout, this model offers more effective control over the details and alignment of the synthesised images with their corresponding semantic layouts.

The SEAN \cite{Zhu_2020_CVPR} model takes this a step further by introducing Semantic Region-Adaptive Normalisation (SEAN), which allows for per-region style control. Building on SPADE's foundation, SEAN enables the use of different style images for each semantic region, resulting in more detailed and realistic image synthesis. The integration of these styles into the normalisation layers allows SEAN to produce high-quality images with fine control over regional styles.

ECGAN \cite{tang2023edge} tackles the challenges of synthesising local details and structures in semantic image synthesis by using edge information as an intermediate representation. This model introduces an attention-guided edge transfer module and a novel contrastive learning method to enhance semantic consistency and detail. By focusing on both local and global semantic information, ECGAN significantly improves the quality of synthesised images, particularly in capturing fine details and small objects that previous models often miss.

SurfelGAN \cite{Yang_2020_CVPR} brings a novel approach to generating realistic sensor data for autonomous driving simulations. By combining texture-mapped surfels, a 3D reconstruction method, with GANs, SurfelGAN creates realistic camera images from novel viewpoints, evaluated on the Waymo Open dataset \cite{Sun_2020_CVPR}. This method vastly improves the realism of synthetic data compared to traditional simulation methods using gaming engines, enhancing the training and testing of ADS.

The OASIS \cite{schonfeld2021you} model simplifies the GAN framework for semantic image synthesis by redesigning the discriminator to use semantic label maps directly as ground truth for training. This innovation eliminates the need for a VGG-based perceptual loss, which is typically used to enhance the quality of synthesised images, thereby simplifying the training process and improving the fidelity of the output images.

Finally, the Robusta \cite{hariat2023learning} model focuses on improving the robustness of semantic segmentation models by generating high-quality, perturbed images using a novel conditional GAN architecture. Robusta utilises a two-stage GAN architecture: a coarse generator for handling label-to-image translation and a fine generator for improving image quality. The inclusion of attention layers and spatially-adaptive normalisation enables better handling of anomalies and distribution shifts, significantly enhancing the robustness of segmentation models in real-world scenarios.

\subsection{Unpaired Image-to-Image Translation Models}
Unpaired I2I translation models are a distinct type of conditional GANs where the input and output in the training dataset do not have a direct correspondence. These models typically require a consistency loss between the input and output domains, along with a GAN loss during training. For example, the pioneering CycleGAN \cite{zhu2020unpaired} introduced the cycle-consistency loss \(\mathcal{L}_{cyc}\), which can be defined as follows:

\begin{equation}
\begin{split}
        \mathcal{L}_{cyc}(G, F) &= \mathbb{E}_{\sensordata \sim p_{data}(\sensordata)} [ \| F(G(\sensordata))
        \\&- \sensordata \|_1] + \mathbb{E}_{\mathbf{y} \sim p_{data}(\mathbf{y})} \left[ \| G(F(\mathbf{y})) - \mathbf{y} \|_1 \right],
\end{split}
\end{equation}
where \( G \) is the generator mapping from domain \( X \) to domain \( Y \), \( F \) is the generator mapping from domain \( Y \) to domain \( X \), \( \sensordata \in X \) is an image from domain \( X \), and \( \mathbf{y} \in Y \) is an image from domain \( Y \).\par
For sensor simulation, these models are typically employed for tasks such as sim-to-real mapping or weather modelling, where creating paired data in different domains is super challenging. The summary of unpaired I2I translation models based on GANs is provided in Table \ref{tab:unpaired_i2i_models} In the following, we will separately review the unpaired I2I models for synthesising Camera RGB images and Lidar point clouds.
\subsubsection{RGB Image synthesis}
Building on CycleGAN's concept, most unpaired I2I models have introduced methods to enforce consistency. DistanceGAN \cite{10.5555/3294771.3294843} maintains structural relationships without inverse mapping, effective for translating different weather conditions. CyCADA \cite{pmlr-v80-hoffman18a} combines pixel-level and feature-level domain adaptation with cycle-consistency constraints, preserving semantic content. MUNIT \cite{huang2018munit} uses content and style codes to improve diversity and quality. The CUT model \cite{10.1007/978-3-030-58545-7_19} employs contrastive learning to enhance translation by maximising mutual information, and simplifying training. SRUNIT \cite{jia2021semantically} introduces a robustness loss for semantic invariance, improving semantic integrity. Spatially-correlative loss \cite{zheng} preserves scene structure while allowing appearance changes, enhancing structural consistency. Jung et al. \cite{jung2022exploring} use semantic relation consistency regularisation and decoupled contrastive learning to improve spatial correspondence and semantic alignment. SPR \cite{xie2023unpaired} finds the shortest path between domains, enhancing content preservation while transforming appearance.
\begin{table*}[!h]
\centering
\caption{Summary of unpaired I2I translation GAN-based models.}
\label{tab:unpaired_i2i_models}
\begin{tblr}{
colspec = {Q[110,c]Q[30,c]Q[20,c]Q[430,l]Q[140,l]},
}
\hline
\textbf{Model} & \textbf{Sensor} & \textbf{Year} & \textbf{Description} & \textbf{Datasets} \\ \hline
CycleGAN \cite{zhu2020unpaired} & Camera & 2017 & Introduces cycle-consistency loss for unpaired image-to-image translation & Cityscapes \cite{Cordts2016Cityscapes} \\ \hline
DistanceGAN \cite{10.5555/3294771.3294843} & Camera & 2017 & Maintains structural relationships within images during translation & Cityscapes \cite{Cordts2016Cityscapes} \\ \hline
CyCADA \cite{pmlr-v80-hoffman18a} & Camera & 2017 & Combines pixel-level and feature-level domain adaptation with cycle-consistency constraints & GTA-V \cite{Richter_2016_ECCV}, SYNTHIA \cite{Ros_2016_CVPR} \\ \hline
MUNIT \cite{huang2018munit} & Camera & 2018 & Decomposes image representation into content and style codes for diverse outputs & Cityscapes \cite{Cordts2016Cityscapes}, SYNTHIA \cite{Ros_2016_CVPR} \\ \hline
Saleh et al. \cite{9022327} & Lidar & 2019 & Employs CycleGAN for sim-to-real mapping on BEV Lidar point clouds & KITTI \cite{Geiger2013IJRR}, CARLA \cite{Dosovitskiy2017CARLAAO} \\ \hline
Sensor Transfer \cite{Carlson2018SensorTL} & Camera & 2018 & Transfers sensor effects from real datasets to synthetic ones to improve robustness & KITTI \cite{Geiger2013IJRR}, GTA-V \cite{Richter_2016_ECCV}, Cityscapes \cite{Cordts2016Cityscapes} \\ \hline
CUT \cite{10.1007/978-3-030-58545-7_19} & Camera & 2020 & Uses contrastive learning to enhance unpaired I2I translation & Cityscapes \cite{Cordts2016Cityscapes} \\ \hline
AnalogicalGAN \cite{Gong2020} & Camera & 2020 & Learns from synthetic images to perform zero-shot image translation for foggy scenes & Cityscapes \cite{Cordts2016Cityscapes}, V-KITTI \cite{gaidon2016virtual}\\ \hline
Tremblay et al. \cite{Tremblay2020} & Camera & 2020 & Integrates multiple methods for realistic rain augmentation on image datasets & KITTI \cite{Geiger2013IJRR}, Cityscapes \cite{Cordts2016Cityscapes}, nuScenes \cite{nuscenes} \\ \hline
SRUNIT \cite{jia2021semantically} & Camera & 2021 & Enforces semantic robustness to address semantics flipping in image translation & Cityscapes \cite{Cordts2016Cityscapes} \\ \hline
Richter et al. \cite{9756256} & Camera & 2021 & Enhances realism of synthetic images by leveraging intermediate representations & GTA-V \cite{Richter_2016_ECCV}, Cityscapes \cite{Cordts2016Cityscapes} \\ \hline
PCT \cite{xiao2021synlidar} & Lidar & 2021 & Decomposes synthetic-to-real gap into appearance and sparsity components & SynLiDAR, Semantic-KITTI \cite{behley2019iccv} \\ \hline
USIS \cite{ESKANDAR202314} & Camera & 2021 & Uses SPADE generator with self-supervised segmentation loss for realistic image synthesis & Cityscapes \cite{Cordts2016Cityscapes} \\ \hline
Jung et al. \cite{jung2022exploring} & Camera & 2022 & Enhances spatial correspondence and semantic alignment with semantic relation consistency & Cityscapes \cite{Cordts2016Cityscapes} \\ \hline
Eskandar et al. \cite{eskandar2023pragmatic} & Camera & 2023 & Uses synthetic semantic layout to generate real RGB images maintaining content integrity & GTA-V \cite{Richter_2016_ECCV}, Cityscapes \cite{Cordts2016Cityscapes}, Mapillary \cite{8237796} \\ \hline
SPR \cite{xie2023unpaired} & Camera & 2023 & Encourages network to find shortest path connecting two domains in unpaired translation & Cityscapes \cite{Cordts2016Cityscapes} \\ \hline
Barrera et al. \cite{10.1109/ITSC48978.2021.9564553} & Lidar & 2023 & Preserves small object details during domain adaptation on BEV point clouds & CARLA \cite{Dosovitskiy2017CARLAAO}, KITTI \cite{Geiger2013IJRR} \\ \hline
CLS2R \cite{haghighi2023contrastive} & Lidar & 2023 & Uses contrastive learning for sim-to-real mapping of Lidar point clouds & CARLA \cite{Dosovitskiy2017CARLAAO}, KITTI \cite{Geiger2013IJRR} \\ \hline
\end{tblr}
\end{table*}


The other models in this domain are miscellaneous, as reviewed in the following paragraphs.
\par
Sensor Transfer \cite{Carlson2018SensorTL} tackles the domain gap between synthetic and real datasets by transferring sensor effects such as chromatic aberration, blur, exposure, noise, and colour temperature from real datasets to synthetic ones. This approach improves the robustness and generalisability of models trained on synthetic data when applied to real-world tasks, enhancing the realism of synthetic data more effectively than traditional domain randomisation techniques.

Richter et al. \cite{9756256} enhance the realism of synthetic images by leveraging intermediate representations produced by conventional rendering pipelines. Their method integrates a convolutional network trained via a novel adversarial objective, aligning scene layout distributions across datasets to reduce artefacts and significantly enhance the stability and realism of synthesised images.
\begin{figure*}[!t]
\centerline{\includegraphics[width=\linewidth]{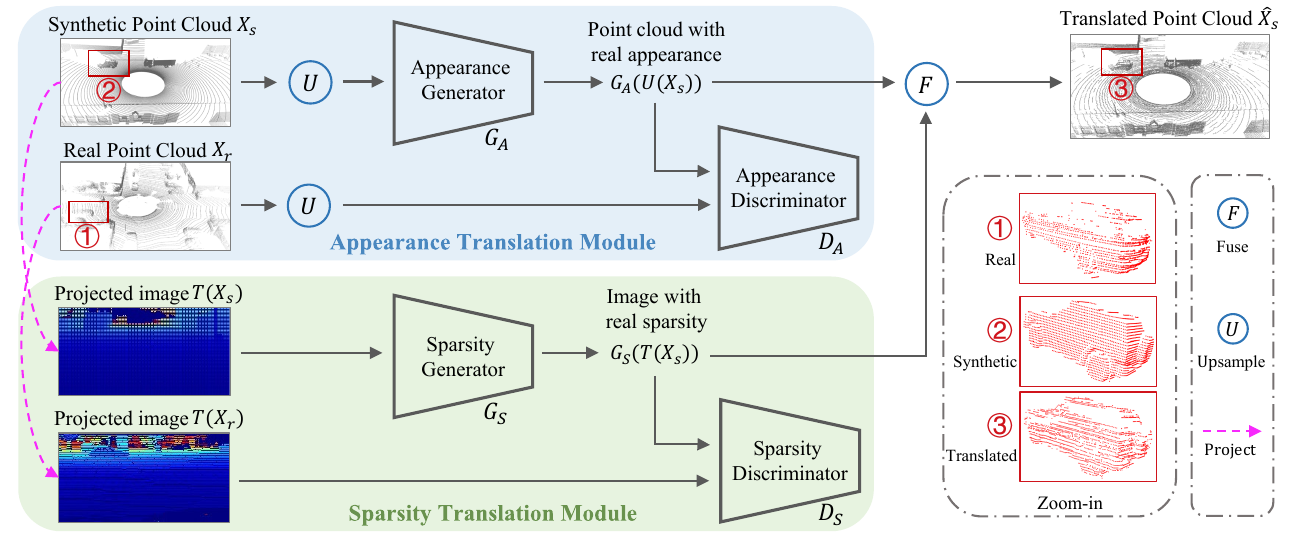}}
\caption{The training data flow of PCT \cite{xiao2021synlidar}, an unpaired data translation model based on GANs for sim-to-real mapping of Lidar point clouds. }
\label{fig:u_i2i_GAN}
\end{figure*}
\begin{table*}
\centering
\caption{Summary of GAN-based video Prediction models (all used for camera RGB image synthesis).}
\label{tab:video_gan}
\begin{tblr}{
colspec = {Q[70]Q[20,c]Q[420,l]Q[120,l]},
}
\hline
\textbf{Model} & \textbf{Year} & \textbf{Description} & \textbf{Datasets} \\ \hline
GeoSim \cite{Chen2021GeoSimRV} & 2021 & Creates realistic video simulations using 3D objects and a physics engine. & Argoverse \cite{Argoverse} \\ \hline
DriveGAN \cite{Kim2021_DriveGAN} & 2021 & Learns from video footage and actions for interactive scene editing and future frame prediction. & CARLA \cite{Dosovitskiy2017CARLAAO}, RWD \cite{Kim2021_DriveGAN} \\ \hline
\end{tblr}
\end{table*}
The USIS \cite{ESKANDAR202314} model marks a significant advance in creating realistic images from segmentation masks without paired data. This framework leverages a SPADE generator enhanced with a self-supervised segmentation loss and a wavelet-based discriminator, ensuring high semantic consistency and detailed textures.

Eskandar et al. \cite{eskandar2023pragmatic} propose a framework for pragmatic semantic image synthesis (SIS) for urban scenes, using synthetic semantic layout to generate real RGB images that maintain the content of the input mask while adopting the appearance of real images.

AnalogicalGAN \cite{Gong2020} introduces the concept of analogical image translation, learning from synthetic clear-weather and foggy images to translate real clear-weather images to real foggy images without seeing any real foggy images during training. This approach allows AnalogicalGAN to perform zero-shot image translation~\cite{gaidon2016virtual}.

Tremblay et al. \cite{Tremblay2020} present a comprehensive rain rendering pipeline designed to improve the robustness of computer vision algorithms under rainy conditions. This approach integrates physics-based, data-driven, and hybrid methods to generate realistic rain effects on existing image datasets. The importance of realistic rain augmentation is highlighted through extensive validation on datasets such as KITTI, Cityscapes, and nuScenes \cite{nuscenes}.

\subsubsection{Lidar Point Cloud Synthesis}
Unpaired I2I models for Lidar point cloud synthesis usually first transform the 3D point cloud into range images which are further processed by image-based models. The first notable model \cite{9022327} employs a CycleGAN-based framework to address the domain shift between synthetic and real Lidar point cloud data, particularly for vehicle detection from a bird’s eye view (BEV). This model enhances previous methods by maintaining the structural integrity of real Lidar data during translation, significantly improving vehicle detection performance.

The ePointDA model \cite{Zhao2020ePointDAAE} presents an end-to-end sim-to-real domain adaptation framework for Lidar point cloud segmentation, bridging the domain gap at both pixel and feature levels. Unlike previous multi-stage pipelines, it includes self-supervised dropout noise rendering, statistics-invariant and spatially-adaptive feature alignment, and transferable segmentation learning. This approach enhances Lidar segmentation by rendering dropout noise for synthetic data and aligning features spatially between domains without relying on real-world statistics.

The Barrera et al. \cite{10.1109/ITSC48978.2021.9564553} build upon the CycleGAN framework by incorporating semantic consistency to preserve small object details during domain adaptation on BEV point clouds. This model excels in maintaining information about small objects, such as pedestrians and cyclists, which are often lost in traditional methods. \par
The PCT (Point Cloud Translator) model \cite{xiao2021synlidar} addresses domain gaps between synthetic and real Lidar point clouds by decomposing the synthetic-to-real gap into appearance and sparsity components. It employs an Appearance Translation Module (ATM) to up-sample synthetic point clouds and translate them to resemble real point clouds, followed by a Sparsity Translation Module (STM) to integrate real sparsity features (see Fig. \ref{fig:u_i2i_GAN}). This innovative approach effectively mitigates domain gaps, resulting in high-quality translations that enhance semantic segmentation tasks.
\par
The CLS2R model \cite{haghighi2023contrastive} introduces a contrastive learning framework for sim-to-real mapping of Lidar point clouds, using a lossless representation of Lidar data that includes depth, reflectance, and raydrop attributes. This model enhances realism and faithfulness by using contrastive learning to ensure high similarity between input and output patches, effectively synthesising realistic Lidar point clouds.
\subsection{Video Prediction Models}
Video prediction models represent another type of conditional model, where the goal is to predict future video frames conditioned on the preceding ones. In opposition to previous GAN categories, these models incorporate the temporal element, requiring them to effectively model the movement of dynamic objects. The summary of video prediction models based on GANs is provided in Table \ref{tab:video_gan}. In the following, we review two notable works in this area, both applied to RGB image synthesis.

GeoSim \cite{Chen2021GeoSimRV} introduces a geometry-aware image composition process for creating realistic video simulations for self-driving applications. This model tackles the challenges of photo-realism and high-level control by utilising a diverse bank of 3D objects derived from sensor data. By proposing realistic object placements, rendering dynamic objects in new poses, and seamlessly blending them into existing scenes, GeoSim ensures traffic-aware, geometrically consistent synthetic images. Furthermore, the model incorporates a physics engine to simulate realistic vehicle dynamics and interactions, significantly enhancing the overall realism of the simulations.
\par
\begin{figure}[!t]
\centerline{\includegraphics[width=\linewidth]{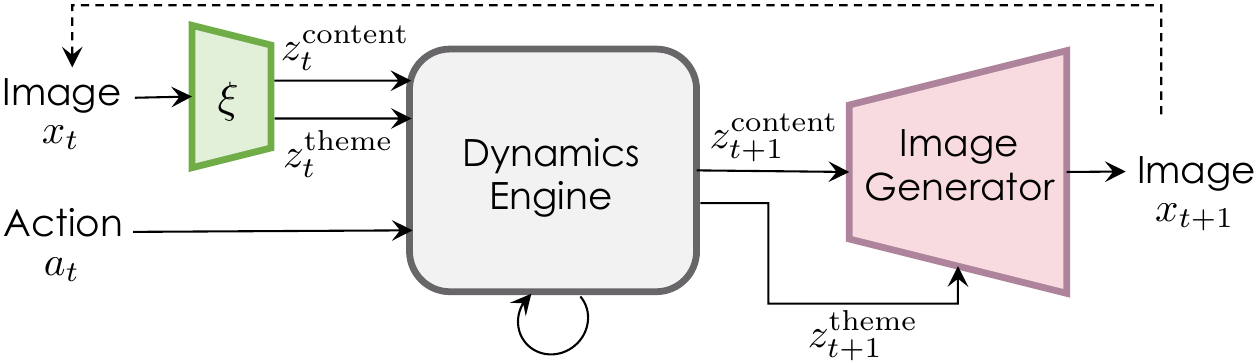}}
\caption{The inference pipeline of DriveGAN \cite{Kim2021_DriveGAN}, a video prediction model based on GANs for RGB images. }
\label{fig:video_gan}
\end{figure}

\begin{table*}[!h]
\centering
\caption{Summary of unconditional diffusion models.}
\label{tab:uc_dm}
\begin{tblr}{
colspec = {Q[80,c]Q[50,c]Q[20,c]Q[440,l]Q[80,l]},
}
\hline
\textbf{Model} & \textbf{Sensor} & \textbf{Year} & \textbf{Description} & \textbf{Datasets} \\ \hline
LiDARGen \cite{zyrianov2022learning} & Lidar & 2022 & Generates point clouds using a stochastic denoising process ensuring physical feasibility. & KITTI-360 \cite{Liao2022PAMI}, NuScenes \cite{nuscenes}\\ \hline
Parke et al. \cite{park2023learning} & Camera & 2023 & Simultaneously generate images and semantic layouts with a combined diffusion model. & Cityscapes \cite{Cordts2016Cityscapes} \\ \hline
R2DM \cite{nakashima2024lidar} & Lidar & 2024 & Uses a diffusion model to generate diverse and high-fidelity 3D scene point clouds. & KITTI-360  \cite{Liao2022PAMI}, KITTI-Raw \cite{Geiger2013IJRR} \\ \hline
RangeLDM \cite{hu2024rangeldm} & Lidar & 2024 & Utilises latent diffusion to generate high-quality range-view Lidar point clouds. & KITTI-360  \cite{Liao2022PAMI}, NuScenes \cite{nuscenes} \\ \hline
LiDM \cite{zyrianov2024lidardm} & Lidar & 2024 & State-of-the-art method for generating realistic Lidar scenes with advanced techniques. & KITTI-360  \cite{Liao2022PAMI}, NuScenes \cite{nuscenes}\\ \hline
\end{tblr}
\end{table*}
DriveGAN \cite{Kim2021_DriveGAN} employs a neural simulator that learns from sequences of video footage and the actions taken by an ego-agent within an environment. Utilising a VAE and GANs, it learns a latent space for images, enabling the dynamics engine to learn transitions within this space (as shown in Fig. \ref{fig:video_gan}). DriveGAN distinguishes itself by disentangling different components of a scene without supervision, allowing users to interactively edit scenes and generate unique scenarios. Additionally, the model features an action-conditioned component that predicts future frames based on the current state and intended actions, offering a robust framework for simulating driving behaviours.
\section{Generative Models -- Diffusion Models} \label{sec:DMs}
Diffusion models have set new benchmarks in various data generation tasks due to their inherent advantages over GAN models, such as more stable training and the ability to iteratively refine samples, resulting in higher quality synthesis. These models have found extensive applications in sensor simulation for autonomous driving, spanning a range of settings from basic unconditional models to text-to-image models and even spatiotemporal video prediction models, as discussed below.
\begin{table*}[!h]
\centering
\caption{Summary of T2I diffusion models (all used for camera RGB image synthesis).}
\label{tab:t2i_dm}
\begin{tblr}{
colspec = {Q[100,c]Q[20,c]Q[420,l]Q[70,l]},
}
\hline
\textbf{Model} & \textbf{Year} & \textbf{Description} & \textbf{Datasets} \\ \hline
SDM \cite{wang2022semantic} & 2022 & Processes semantic layouts and noisy images separately, embedding the layout into the denoising network. & Cityscapes \cite{Cordts2016Cityscapes} \\ \hline
GEODIFFUSION \cite{chen2024geodiffusion} & 2023 & Proposes a text-prompted geometric control framework for high-quality object detection data. & NuScenes \cite{nuscenes} \\ \hline
BEVControl \cite{yang2023bevcontrol} & 2023 & Generates realistic and controllable street-view images from BEV sketches. & NuScenes \cite{nuscenes} \\ \hline
MAGICDRIVE \cite{gao2023magicdrive} & 2023 & Uses 3D geometry controls like camera poses, road maps, and bounding boxes with textual descriptions. & NuScenes \cite{nuscenes} \\ \hline
Loiseau et al. \cite{loiseau2023reliability} & 2023 & Generates synthetic data to assess perception model reliability under domain shifts. & Cityscapes \cite{Cordts2016Cityscapes} \\ \hline
Text2Street \cite{su2024text2street} & 2024 & Framework for generating controllable street-view images from text, including road topology and weather. & NuScenes \cite{nuscenes} \\ \hline
\end{tblr}
\end{table*}
\subsection{Unconditional Models}
Unconditional diffusion models generate data from a simple prior distribution, such as Gaussian noise, and iteratively refine it through a series of transformations driven by the diffusion process (as discussed in Section \ref{sec:background}). The summary of unconditional DMs is provided in Table \ref{tab:uc_dm}. The following sections separately review the unconditional diffusion models for both camera image synthesis and Lidar point cloud generation.
\subsubsection{RGB Image Synthesis}
In the area of unconditional diffusion models for RGB images, we identified a single notable work. The Gaussian-categorical diffusion process \cite{park2023learning} model presents an RGB image synthesis model by simultaneously generating images and their corresponding semantic layouts. This technique enhances image quality by incorporating semantic understanding into the generation process, using a combined Gaussian and categorical diffusion model to represent the joint distribution of image-layout pairs. 
\subsubsection{Lidar Point Cloud Synthesis}
The pioneering approach that uses diffusion models for unconditional Lidar point cloud generation is LiDARGen \cite{zyrianov2022learning}. This model formulates the point cloud generation process as a stochastic denoising process in the equirectangular view. LiDARGen advances previous methods by ensuring the physical feasibility and controllability of generated samples. Its capability to sample point clouds conditioned on inputs without retraining makes LiDARGen particularly useful for Lidar densification tasks in ADS.

Building on this, the R2DM \cite{nakashima2024lidar} introduces a DM-based framework for Lidar data. R2DM generates diverse and high-fidelity 3D scene point clouds based on the image representation of range and reflectance intensity. This model focuses on stable training, sample quality, and versatility in handling inverse problems. The R2DM model has demonstrated remarkable performance on the KITTI-360 and KITTI-Raw datasets, particularly excelling in Lidar completion tasks.

RangeLDM \cite{hu2024rangeldm} leverages latent diffusion models to rapidly generate high-quality range-view Lidar point clouds. This model projects point clouds onto range images using Hough Voting for accurate range-view data distribution. The range images are then compressed into a latent space with a VAE and processed by a diffusion model to enhance expressivity.

Finally, the LiDM \cite{zyrianov2024lidardm} model presents a SOTA approach for generating realistic Lidar scenes using diffusion models. LiDM focuses on pattern realism, geometry realism, and object realism by introducing three core innovations: curve-wise compression, point-wise coordinate supervision, and patch-wise encoding. These techniques ensure that the generated Lidar scenes maintain high fidelity to real-world data, preserving the detailed geometric and structural properties of objects and their surroundings.

\subsection{Text-to-Image Translation Models}
Building on unconditional DMs, several methods have incorporated auxiliary data for enhanced control and synthesis quality. For instance, the Semantic Diffusion Model (SDM) \cite{wang2022semantic} processes semantic layouts and noisy images separately, embedding the semantic layout into the decoder of the denoising network. The reverse diffusion process for this model can be described as:
\begin{equation}
    q(\mathbf{y}_{t-1} | \mathbf{y}_t, \sensordata) = \mathcal{N}(\mathbf{y}_{t-1}; \mu_\theta(\mathbf{y}_t, t, \sensordata), \Sigma_\theta(\mathbf{y}_t, t, \sensordata)).
\end{equation}
where \( \sensordata \) represents the condition, such as a semantic layout or text prompt, and \( \mathbf{y} \) is the image being modelled. The summary of T2I translation DMs is provided in Table \ref{tab:t2i_dm}. The following sections will review other conditional diffusion models, all of which have been applied to RGB image synthesis.
\par
\begin{figure}[!t]
\centerline{\includegraphics[width=\linewidth]{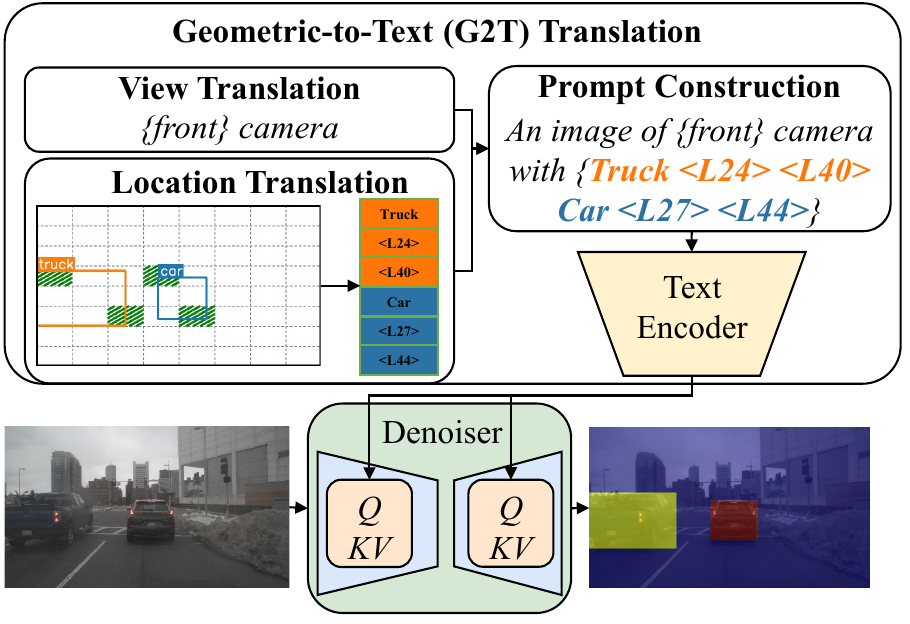}}
\caption{The model architecture of GeoDiffusion \cite{chen2024geodiffusion}, a T2I translation diffusion model for RGB images. }
\label{fig:t2i_dm}
\end{figure}

Expanding on SDM's advancements, GEODIFFUSION \cite{chen2024geodiffusion} proposes a text-prompted geometric control framework for generating high-quality object detection data. This method leverages pre-trained T2I diffusion models to translate various geometric conditions into text prompts, as shown in Fig. \ref{fig:t2i_dm}. In contrast to previous layout-to-image methods that encoded only bounding boxes, GEODIFFUSION incorporates additional geometric conditions such as camera views, enhancing the realism and consistency of generated images.
\par
Further refining this concept, BEVControl \cite{yang2023bevcontrol} introduces a robust approach for generating realistic and controllable street-view images from BEV sketches. By incorporating a two-stage framework that decouples visual consistency into geometry and appearance sub-goals, BEVControl allows for detailed manipulation of background and foreground elements.
\begin{table*}[!h]
\centering
\caption{Summary of diffusion video prediction models.}
\label{tab:video_dm}
\begin{tblr}{
colspec = {Q[120,c]Q[50,c]Q[20,c]Q[350,l]Q[80,l]},
}
\hline
\textbf{Model} & \textbf{Sensor} & \textbf{Year} & \textbf{Description} & \textbf{Datasets} \\ \hline
DriveDreamer \cite{wang2023drivedreamer} & Camera & 2023 & Constructs world models from driving videos and driver behaviours, enhancing sampling efficiency with traffic structural information. & NuScenes \cite{nuscenes} \\ \hline
Drive-WM \cite{wang2023driving} & Camera & 2023 & Features joint spatial-temporal modelling for generating high-fidelity multi-view videos in driving scenes. & NuScenes \cite{nuscenes} \\ \hline
DrivingDiffusion \cite{li2023drivingdiffusion} & Camera & 2023 & Introduces a spatial-temporal consistent diffusion framework for generating realistic multi-view videos controlled by 3D layout. & NuScenes \cite{nuscenes} \\ \hline
Panacea \cite{wen2023panacea} & Camera & 2023 & Generates panoramic and controllable videos for autonomous driving scenarios with a two-stage system and 4D attention mechanism. & NuScenes \cite{nuscenes} \\ \hline
ADriver-I \cite{jia2023adriveri} & Camera & 2023 & Unifies control signal prediction and future scene generation within a single framework using a multimodal large language model. & NuScenes \cite{nuscenes}\\ \hline
WoVoGen \cite{lu2023wovogen} & Camera & 2023 & Combines an explicit 4D world volume with diffusion models to generate high-quality, multi-camera street-view videos. & NuScenes \cite{nuscenes}\\ \hline
SubjectDrive \cite{huang2024subjectdrive} & Camera & 2024 & Enhances the scalability and diversity of generative data by integrating external subjects into the generation process. & NuScenes \cite{nuscenes}\\ \hline
DriveDreamer-2 \cite{zhao2024drive} & Camera & 2024 & Enhances DriveDreamer with a Large Language Model to generate user-defined driving videos from text prompts. & NuScenes \cite{nuscenes}\\ \hline
LidarDM \cite{ran2024towards} & Lidar & 2024 & Produces realistic, layout-aware, and temporally coherent Lidar point cloud videos, useful for autonomous driving simulations. & KITTI~\cite{Geiger2013IJRR}, NuScenes \cite{nuscenes}\\ \hline
\end{tblr}
\end{table*}
\par
Taking this approach further, MAGICDRIVE \cite{gao2023magicdrive} incorporates diverse 3D geometry controls, such as camera poses, road maps, and 3D bounding boxes, along with textual descriptions. By using a cross-view attention module, MAGICDRIVE ensures consistency across multiple camera views, significantly enhancing the realism and geometric accuracy of generated scenes. Its ability to control various scene attributes, such as weather conditions and time of day, is particularly useful for BEV segmentation and 3D object detection tasks.
\par
Loiseau et al. \cite{loiseau2023reliability} introduce an innovative approach to generating realistic synthetic data for assessing the reliability of perception models under various domain shifts. By leveraging a pre-trained T2I diffusion model, specifically Stable Diffusion \cite{rombach2021highresolution}, augmented with a ControlNet \cite{zhang2023adding} module, this method conditions generation on semantic masks from the Cityscapes dataset while using text prompts to simulate different target domains. This approach enables zero-shot generation of synthetic data that aligns with the semantic conditions of the original domain while reflecting the visual properties of the target domains.
\par
Finally, Text2Street \cite{su2024text2street} offers a framework for generating controllable street-view images from textual descriptions. Introducing three main components—a lane-aware road topology generator, a position-based object layout generator, and a multiple control image generator—Text2Street generates detailed and semantically accurate street-view images based on specific textual descriptions of road topology, traffic status, and weather conditions.

\subsection{Video Prediction Models}
Video prediction diffusion models (also referred to as world models) are a form of conditional diffusion models where the input consists of previous frames and the goal is to predict future ones. These models differ from standard conditional diffusion models by needing to model temporal interactions and being more computationally efficient due to higher input dimensionality. They can also incorporate various auxiliary conditions, such as actions, bounding box layouts, and text prompts, to enhance controllability. The summary of video prediction DMs is provided in Table \ref{tab:video_dm}. The following sections review several pioneering works in this domain for both RGB image and Lidar point cloud synthesis.

\subsubsection{RGB Image Synthesis}

\begin{figure}[!t]
\centerline{\includegraphics[width=\linewidth]{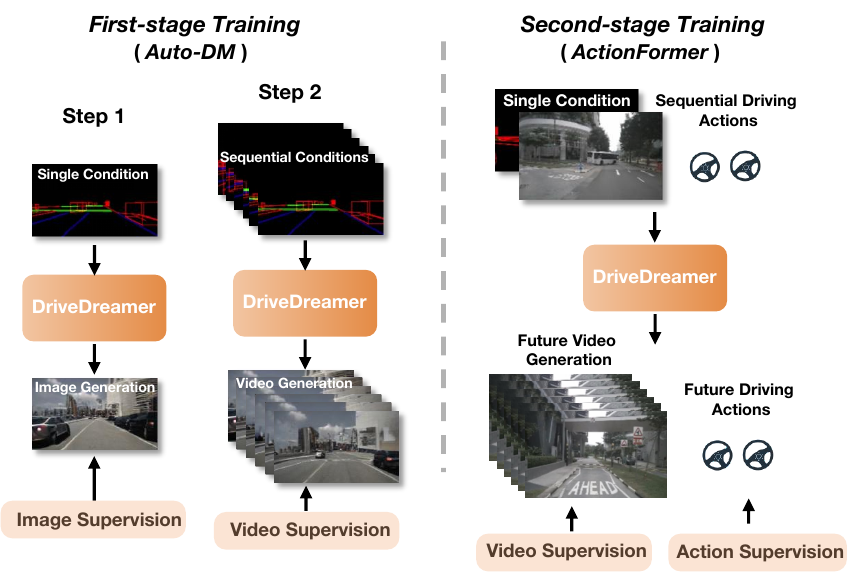}}
\caption{The two-stage training pipeline of DriveDreamer \cite{wang2023drivedreamer}, a video prediction diffusion model for RGB images. }
\label{fig:video_dm}
\end{figure}

DriveDreamer \cite{wang2023drivedreamer} is a pioneering model that constructs comprehensive world models from real-world driving videos and human driver behaviours. It introduces the Autonomous-driving Diffusion Model (Auto-DM) with a two-stage training pipeline, as shown in Fig. \ref{fig:video_dm}. The first stage enhances sampling efficiency by incorporating traffic structural information, while the second stage focuses on video prediction to establish the world model. This approach generates high-quality, controllable driving videos and reasonable driving policies.

Building on this foundation, Drive-WM \cite{wang2023driving} is the first driving world model compatible with existing end-to-end planning models. It features joint spatial-temporal modelling facilitated by view factorisation, enabling the generation of high-fidelity multi-view videos in driving scenes. Drive-WM predicts multiple futures based on distinct driving manoeuvres and determines the optimal trajectory using image-based rewards. Drive-WM leverages latent video diffusion models for multi-view and temporal modelling, ensuring consistency and quality across different views and frames.

\par
DrivingDiffusion \cite{li2023drivingdiffusion} introduces a spatial-temporal consistent diffusion framework for generating realistic multi-view videos controlled by 3D layout. The method addresses three main challenges: cross-view consistency, cross-frame consistency, and instance quality. The multi-stage scheme includes a multi-view single-frame image generation model, a single-view video generation step shared by multiple cameras, and post-processing to enhance video length and consistency. The use of information exchange between adjacent cameras and a temporal sliding window algorithm ensures the generation of high-quality, realistic driving videos. 
\begin{table*}[!h]
\centering
\caption{Summary of miscellaneous generative models.}
\label{tab:generative_misc}
\begin{tblr}{
colspec = {Q[90,c]Q[100,c]Q[20,c]Q[350,l]Q[100,l]},
}
\hline
\textbf{Model} & \textbf{Category} & \textbf{Year} & \textbf{Description} & \textbf{Datasets} \\ \hline
CRN \cite{Chen2017PhotographicIS} & Supervised I2I & 2017 & Synthesises photographic images from semantic layouts using cascaded refinement modules, without adversarial training. & Cityscapes \cite{Cordts2016Cityscapes}\\ \hline
Vecek et al. \cite{9339933} & Supervised I2I & 2020 & Predicts the strength of Lidar responses using deep learning, enhancing data realism and improving segmentation accuracy. & GTA V \cite{Richter_2016_ECCV}, semantic-KITTI \cite{behley2019iccv}\\ \hline
LiDARsim \cite{manivasagam2020lidarsim} & Supervised I2I & 2020 & Generates realistic Lidar point clouds by leveraging real-world data, combining physics-based simulation and deep neural network refinement. & KITTI \cite{Geiger2013IJRR}, NuScenes \cite{nuscenes}\\ \hline
Hu et al. \cite{10.1007/978-3-030-58517-4_45} & AR Video Prediction & 2020 & Predicts ego-motion, static scenes, and dynamic agent motion probabilistically using a spatio-temporal convolutional module. & Cityscapes \cite{Cordts2016Cityscapes}\\ \hline
RINet \cite{Guillard2022} & Supervised I2I & 2022 & Simulates Lidar sensors by mapping RGB images to corresponding Lidar features, significantly improving Lidar data realism. & CARLA \cite{Dosovitskiy2017CARLAAO}, Waymo \cite{Sun_2020_CVPR}, semantic-KITTI \cite{behley2019iccv} \\ \hline
READ \cite{li2022read} & Supervised I2I & 2022 &Uses neural rendering to generate realistic driving scenes from sparse point clouds with multi-sampling for coherence and efficiency. & KITTI \cite{Geiger2013IJRR}, Brno Urban \cite{ligocki2020brno}\\ \hline
UltraLiDAR \cite{xiong2023learning} & Unconditional AR & 2023 & Uses a VQ-VAE framework to learn compact, discrete 3D representations of scene-level Lidar point clouds, enabling various downstream applications. & Pandaset \cite{10.1109/ITSC48978.2021.9565009}, KITTI-360 \cite{Liao2022PAMI} \\ \hline
MUVO \cite{bogdoll2023muvo} & AR Video Prediction & 2023 & Integrates raw camera and Lidar data to learn a sensor-agnostic geometric world representation, enhancing prediction quality. & CARLA \cite{Dosovitskiy2017CARLAAO} \\ \hline
ETSSR \cite{10173712} & Supervised I2I & 2023 & Accelerates stereo image simulation using Stereo Super Resolution (SSR) with a transformer-based model inspired by Swin Transformer. & CARLA \cite{Dosovitskiy2017CARLAAO} \\ \hline
WorldDreamer \cite{wang2023world} & AR Video Prediction & 2024 & Uses Spatial Temporal Patchwise Transformer (STPT) for visual token prediction, integrating multi-modal prompts for video generation. & NuScenes \cite{nuscenes} \\ \hline
LidarGRIT \cite{haghighi2024taming} & Unconditional AR & 2024 & Addresses limitations in generating realistic raydrop noise with a generative range image transformer model using AR transformer and VQ-VAE. & KITTI-360 \cite{Liao2022PAMI}, KITTI odometry \cite{Geiger2013IJRR}\\ \hline
BEVGen \cite{swerdlow2024streetview} & Supervised I2I & 2024 & Synthesises spatially consistent street-view images conditioned on BEV layouts using VQ-VAE auto-encoders and a causal transformer. & nuScenes \cite{nuscenes}, Argoverse 2 \cite{NEURIPS_DATASETS_AND_BENCHMARKS2021_4734ba6f} \\ \hline
\end{tblr}
\end{table*}
Panacea \cite{wen2023panacea} introduces a novel approach for generating panoramic and controllable videos tailored for autonomous driving scenarios. This method addresses key challenges in video generation: ensuring consistency and controllability. Panacea operates through a two-stage system where the first stage generates multi-view driving scene images, and the second stage extends these images temporally to create video sequences. It incorporates a novel 4D attention mechanism and the ControlNet framework for detailed control using BEV layouts.

ADriver-I \cite{jia2023adriveri} proposes an innovative method that unifies control signal prediction and future scene generation within a single framework. It uses interleaved vision-action pairs as inputs and employs a multimodal large language model (MLLM) and video DM (VDM) to iteratively predict control signals and future frames. Unlike previous methods, ADriver-I does not require extensive prior information such as 3D bounding boxes or HD maps, making it more flexible and generalisable.

WoVoGen \cite{lu2023wovogen} combines an explicit 4D world volume with diffusion models to generate high-quality, multi-camera street-view videos. This model operates in two phases: envisioning a 4D temporal world volume based on vehicle control sequences and generating multi-camera videos informed by this envisioned world volume. By incorporating a dense voxel volume that encapsulates comprehensive data about the scene, WoVoGen ensures both intra-world consistency and inter-sensor coherence.

SubjectDrive \cite{huang2024subjectdrive} presents an innovative video generation framework aimed at enhancing the scalability and diversity of generative data. It integrates external subjects into the generation process via a subject control mechanism, significantly boosting data diversity and utility. The architecture comprises three modules: the Subject Prompt Adapter (SPA) for enriching text embeddings, the Subject Visual Adapter (SVA) for incorporating spatial information, and Augmented Temporal Attention (ATA) for ensuring temporal consistency. 

Building on this progress, DriveDreamer-2 \cite{zhao2024drive} enhances the DriveDreamer framework by incorporating an LLM to generate user-defined driving videos from text prompts. The method separates traffic simulation into foreground (agent trajectories) and background (HDMaps) conditions, using a functional library to finetune the LLM for trajectory generation. DriveDreamer-2 introduces the Unified Multi-View Model (UniMVM) to ensure temporal and spatial coherence in the generated videos, significantly improving video quality.

\subsubsection{Lidar Point Cloud Synthesis}
In the field of Lidar point cloud video prediction, a notable approach is LidarDM \cite{ran2024towards}. This model introduces a generative method capable of producing realistic, layout-aware, and temporally coherent Lidar point cloud videos. In contrast to previous models, LidarDM incorporates driving scenario guidance, making it particularly useful for autonomous driving simulations. It uses a conditional generative framework to ensure the generated Lidar data is physically plausible and aligns with the expected layout of driving environments.

\section{Generative Models -- Miscellaneous} \label{sec:Misc}
Other generative models used for camera and Lidar simulation are often disintegrated into distinct categories. The summary of miscellaneous generative models is provided in Table \ref{tab:generative_misc}. In the following, we review these models in three sub-categories: unconditional AR models, supervised I2I translation models, and AR video prediction models.

\subsection{Unconditional Auto-regressive Models}
Unconditional AR models generate data sequentially without relying on any conditional input, predicting the next value in a sequence based on previously generated values, as discussed in Section \ref{sec:background}. Two notable models in this category, used for Lidar point cloud generation, are UltraLiDAR \cite{xiong2023learning} and LidarGRIT \cite{haghighi2024taming}.

UltraLiDAR \cite{xiong2023learning} builds upon the concept of learning compact, discrete 3D representations of scene-level Lidar point clouds, enabling various downstream applications including unconditional Lidar generation. UltraLiDAR focuses on learning a discrete codebook that captures the geometric structure of scenes and aligns sparse point clouds with dense ones. This model employs a VQ-VAE \cite{NIPS2017_7a98af17} framework and introduces techniques to handle voxelised BEV representation, enhancing computational efficiency.

\par
\begin{figure}[!t]
\centerline{\includegraphics[width=\linewidth]{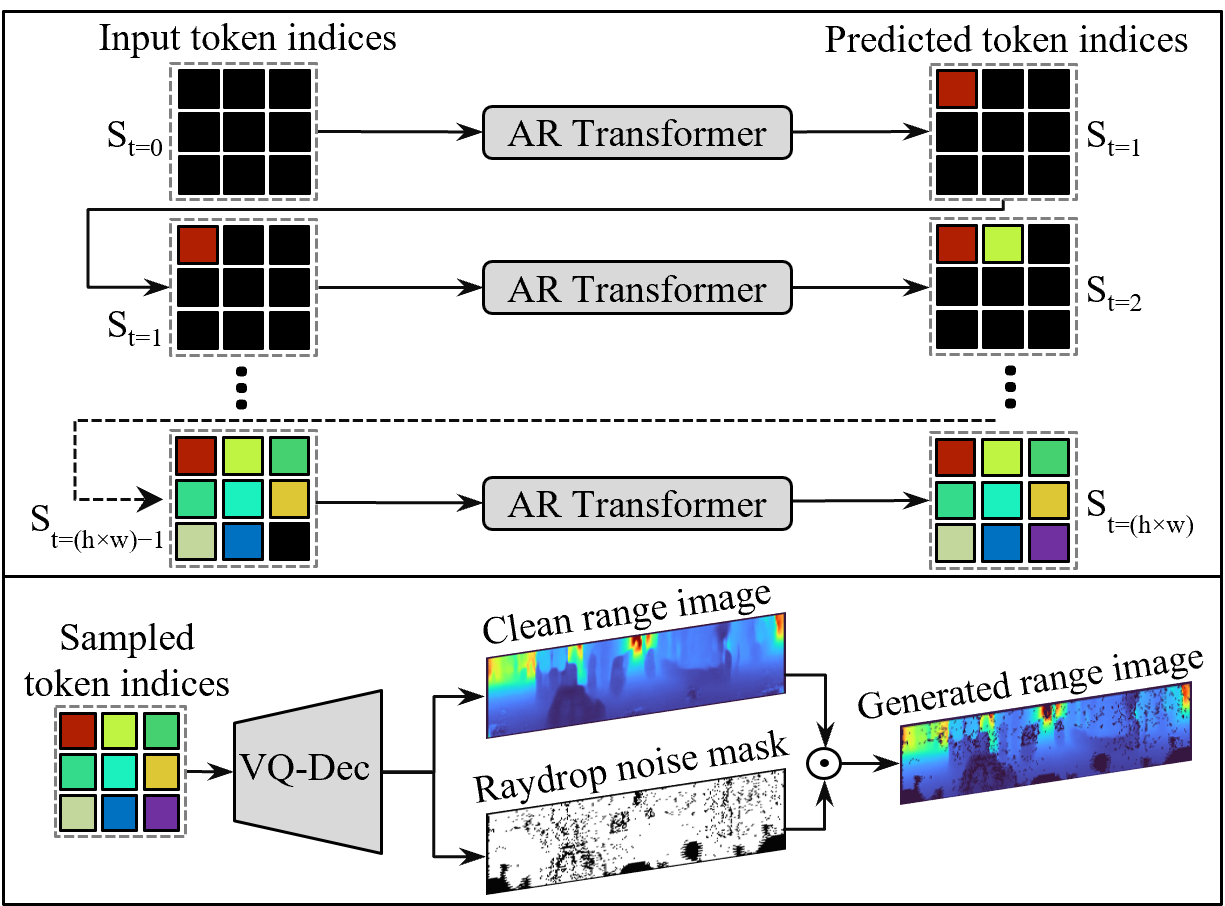}}
\caption{Inference process of the unconditional auto-regressive model, LidarGRIT \cite{haghighi2024taming}, for Lidar point cloud generation. }
\label{fig:uc_ar}
\end{figure}

LidarGRIT \cite{haghighi2024taming} introduces the Lidar generative range image transformer model, which addresses previous DMs' limitations in generating realistic raydrop noise. LidarGRIT uses progressive generation and accurate raydrop noise synthesis through iterative sampling in the latent space via an AR transformer. The sampled tokens are then decoded to range images using an adapted VQ-VAE (see Fig. \ref{fig:uc_ar}). By separating the generation of range images from raydrop noise masks, this method significantly enhances data realism. 
\subsection{Supervised Image-to-Image Translation Models}
Several miscellaneous models can be classified as supervised I2I models. These models do not rely on adversarial, denoising diffusion, or AR prediction losses; instead, they primarily leverage supervised error losses such as perceptual losses or L1/L2 error losses. In the following sections, we review these models, categorising them into either Camera RGB image synthesis or Lidar point cloud synthesis.
\subsubsection{RGB Image Synthesis}
CRN model \cite{Chen2017PhotographicIS} introduces a feedforward network approach to synthesise photographic images from semantic layouts without the need for adversarial training. This model uses cascaded refinement modules to process input semantic layouts through multiple stages, progressively enhancing the resolution and detail of the output image. Each module refines the image by combining downsampled input layouts with upsampled features from the previous module.

READ model \cite{li2022read} introduces a novel neural rendering approach to synthesise photo-realistic driving scenes from sparse point clouds. This method leverages a neural network, which filters and fuses features across different scales to produce detailed and coherent images. In contrast to previous models that struggled with coherence and computational efficiency, READ employs multi-sampling strategies such as Monte Carlo sampling and patch sampling to optimise performance and reduce computational costs.
\begin{figure*}[!t]
\centerline{\includegraphics[width=0.8\linewidth]{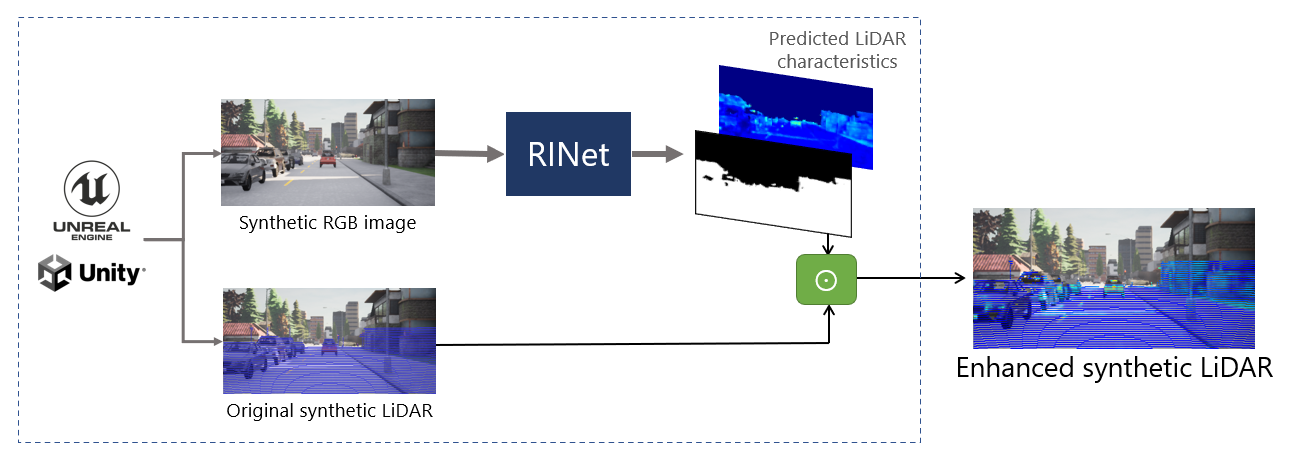}}
\caption{Inference pipeline of RINet \cite{Guillard2022}, a supervised I2I model for Lidar raydrop synthesis. }
\label{fig:supervised_i2i}
\end{figure*}
ETSSR model \cite{10173712} presents a novel method to accelerate stereo image simulation by employing stereo super resolution. The proposed technique initially simulates low-resolution stereo images, then super-resolves them to high-resolution using a novel transformer-based SSR model inspired by Swin Transformer \cite{liu2021Swin}. This method leverages a Disparity-aware Swin Cross Attention Module (DSCAM) for efficient cross-view feature extraction, enhancing both the realism and computational efficiency of the generated images.

BEVGen model \cite{swerdlow2024streetview} employs a novel generative model to synthesise spatially consistent street-view images conditioned on BEV layouts. The model integrates two VQ-VAE auto-encoders for images and BEV representations, and a causal transformer to model high-level scenes. The unique aspect of BEVGen is its ability to handle cross-modal inductive 3D bias, which enables the model to relate information between modalities and across different views. This approach achieves high-quality synthesis results and offers practical applications such as data augmentation and simulated driving scene rendering.
\subsubsection{Lidar Point Cloud Synthesis}
Vacek et al. \cite{9339933} introduce a data-driven method for simulating Lidar sensors by predicting the strength of Lidar responses. This model uses computer-generated data to extract geometrically simulated point clouds and predicts Lidar intensities using deep learning. LPI enhances data realism by accounting for systematic failures and noise, such as low responses on polished surfaces and strong responses on reflective surfaces.

LiDARsim \cite{manivasagam2020lidarsim} presents a novel approach for generating realistic Lidar point clouds by leveraging real-world data. This method involves creating a large catalogue of 3D static maps and dynamic objects by driving around several cities with a self-driving fleet. These assets are then used to compose scenes where a physics-based simulator first performs ray casting over the 3D scene, and a deep neural network refines the output to produce realistic Lidar point clouds. This hybrid approach of combining physics-based and learning-based simulations allows LiDARsim to capture more complex interactions and sensor noise.

Guillard et al. \cite{Guillard2022} introduce a data-driven pipeline to simulate Lidar sensors by mapping RGB images to corresponding Lidar features, such as raydrop and per-point intensities, using real datasets. The Raydrop and Intensity Network (RINet) predicts realistic Lidar characteristics from RGB images, enhancing ray-casted point clouds from standard simulation software (see Fig. \ref{fig:supervised_i2i}). This approach encodes effects such as dropped points on transparent surfaces and high-intensity returns on reflective materials, significantly improving Lidar data realism.
\begin{table*}[!h]
\centering
\caption{Summary of frequency-based NeRFs (all used for camera RGB image synthesis).}
\label{tab:frequency_based_nerf_models}
\begin{tblr}{
colspec = {Q[80,c]Q[20,c]Q[400,l]Q[80,l]},
}
\hline
\textbf{Model} & \textbf{Year} & \textbf{Description} & \textbf{Datasets} \\ \hline
NSG \cite{Ost_2021_CVPR} & 2020 & Decomposes dynamic scenes into scene graphs, representing each object as nodes within a hierarchical structure, enabling efficient rendering of dynamic scenes. & KITTI \cite{Geiger2013IJRR} \\ \hline
Block-NeRF \cite{tancik2022blocknerf} & 2022 & Divides the environment into multiple compact NeRFs, each trained independently, allowing efficient updates without retraining the entire model. & Alamo Square \cite{tancik2022blocknerf} \\ \hline
MapNeRF \cite{wu2023mapnerf} & 2023 & Incorporates map priors into neural radiance fields to synthesise out-of-trajectory driving views with semantic road consistency. & Argoverse2 \cite{NEURIPS_DATASETS_AND_BENCHMARKS2021_4734ba6f} \\ \hline
UC-NeRF \cite{cheng2023uc} & 2023 & Introduces layer-based colour correction, virtual warping, and spatiotemporally constrained pose refinement for high-quality neural rendering with multiple cameras. & Waymo \cite{Sun_2020_CVPR}, NuScenes \cite{nuscenes} \\ \hline
ChatSim \cite{wei2024editable} & 2024 & Enables editable photo-realistic 3D driving scene simulations via natural language commands with external digital assets, using McNeRF and McLight methods. & Waymo \cite{Sun_2020_CVPR} \\ \hline
\end{tblr}
\end{table*}

\subsection{Auto-regressive Video Prediction Models}

There are several video prediction models (i.e. world models) that are primarily based on auto-regressive approaches, utilising spatio-temporal modelling with RNN/GRU models or causal transformers. In the following sections, we will review papers in this domain, all of which have been applied to camera RGB image synthesis or joint RGB image and Lidar point cloud synthesis.

\par
Hu et al. \cite{10.1007/978-3-030-58517-4_45} introduce a deep learning architecture designed for probabilistic future prediction from video data. This model is the first to jointly predict ego-motion, static scenes, and dynamic agent motion in a probabilistic manner. It leverages a spatio-temporal convolutional module, including ConvGRU, to learn representations from RGB video, which can be decoded into future semantic segmentation, depth, and optical flow. The model employs a conditional variational approach to minimise the divergence between present and future distributions, enabling the generation of diverse and accurate future scenarios.

MUVO \cite{bogdoll2023muvo} introduces a multi-modal generative world model with geometric voxel representations, integrating raw camera and Lidar data to learn a sensor-agnostic geometric world representation. In contrast to previous models that focused solely on sensor data, MUVO enhances prediction quality by incorporating 3D occupancy predictions conditioned on actions, showing improvements in both camera and Lidar data prediction.

WorldDreamer \cite{wang2023world} advances the concept of world models for video generation by framing the task as a visual token prediction challenge. This model utilises the Spatial-Temporal Patchwise Transformer (STPT) to focus attention on localised patches within a temporal-spatial window. WorldDreamer excels in generating videos across different scenarios, including natural scenes and driving environments.

\section{Volume Renderers -- NeRFs} \label{sec:NeRFs}
Neural Radiance Fields (NeRFs) \cite{10.1145/3503250} have recently gained significant attention for scene representation and sensor simulation in ADS due to their exceptional ability to produce high-quality novel view synthesis. In contrast to the black-box nature of generative models, NeRFs provide a more transparent and explicit modelling of scenes and radiance. As discussed in Section \ref{sec:NeRFs}, NeRFs use MLP to model the colour and density of a 3D position \(\sensordata \in \mathbb{R}^3\) and the direction of a 2D ray \(\mathbf{d} \in \mathbb{R}^2\). The encoding of this 5D input is crucial for the performance of NeRFs. In the context of ADS, various encoding methods, such as frequency-based, hash-grids, point-based, and multi-planar, are utilised, as discussed in the following sections.

\subsection{Frequency-based Encoding}
Frequency-based encoding is the first technique used for positional encoding and has proven highly effective in capturing and rendering high-frequency details in 3D scenes. This encoding transforms the input 3D coordinates $\mathbf{x}$ and viewing direction $\mathbf{d}$ into a higher-dimensional space using high-frequency functions. The frequency-based encoding function $\gamma$ can be defined as follows:
\begin{align}
        &\gamma(\mathbf{p}) = \big( \sin(2^0 \pi \mathbf{p}), \cos(2^0 \pi \mathbf{p}), \\ &\sin(2^1 \pi \mathbf{p}), \cos(2^1 \pi \mathbf{p}), \ldots, \sin(2^{L-1} \pi \mathbf{p}), \cos(2^{L-1} \pi \mathbf{p}) \big) \nonumber,
\end{align}
where $\mathbf{p} \in \mathbb{R}^3$ can be either the 3D position $\mathbf{x}$ or the 2D viewing direction $\mathbf{d}$, and $L$ is the number of frequency bands used in the encoding. The summary of frequency-based NeRFs is provided in Table \ref{tab:frequency_based_nerf_models}. In the following, we review NeRFs based on frequency-based encoding, which are all used for RGB image synthesis in ADS.
\par
NSG model \cite{Ost_2021_CVPR} introduces a novel neural rendering method that decomposes dynamic scenes into scene graphs, representing each object as nodes within a hierarchical structure. This approach enables efficient and accurate rendering of dynamic scenes by leveraging individual neural radiance fields for each object, in contrast to previous methods that encoded the entire scene into a single network. 
\begin{figure}[!t]
\centerline{\includegraphics[width=\linewidth]{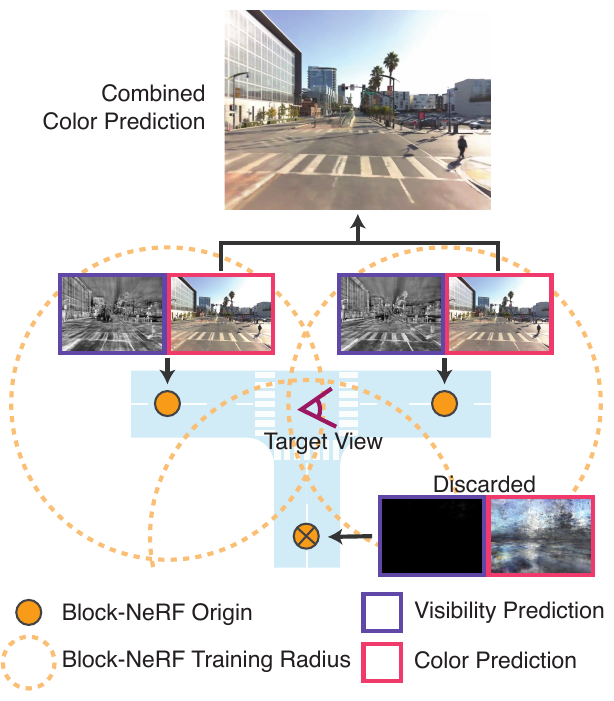}}
\caption{Scene split in Block-NeRF \cite{tancik2022blocknerf}, a volume renderer using frequency-based NeRFs for RGB image synthesis. }
\label{fig:freq_nerf}
\end{figure}
\par
Block-NeRF \cite{tancik2022blocknerf} extends NeRFs to large-scale scenes by dividing the environment into multiple compact NeRFs, each trained independently to cover a specific block of the environment (see Fig. \ref{fig:freq_nerf}). This method addresses the scalability issue of NeRFs, allowing for efficient updates to individual blocks without retraining the entire model. Key innovations include using appearance embeddings, learned pose refinement and controllable exposure to handle data captured under varying environmental conditions. 
\par
Further enhancing the concept of scene representation, MapNeRF \cite{wu2023mapnerf} introduces the incorporation of map priors into neural radiance fields to synthesise out-of-trajectory driving views with semantic road consistency. This method leverages ground and lane information from maps to supervise the density field and warp depth, ensuring multi-view consistency even when the camera deviates from the standard trajectory. 
\par
Addressing the challenges in multi-camera systems, UC-NeRF \cite{cheng2023uc} introduces a method for high-quality neural rendering with multiple cameras in autonomous driving. The key contributions include layer-based colour correction, virtual warping to generate diverse but consistent virtual views, and spatiotemporally constrained pose refinement for robust pose calibration. This approach significantly improves the rendering quality and accuracy of synthesised novel views.
\par
Finally, ChatSim \cite{wei2024editable} presents a novel system enabling editable photo-realistic 3D driving scene simulations via natural language commands with external digital assets. This method leverages a collaborative framework of LLM agents to decouple complex simulation tasks into specific editing operations. The key contributions include McNeRF, a multi-camera neural radiance field method, and McLight, a multi-camera lighting estimation method for the scene-consistent rendering of external digital assets.
\begin{table*}[!h]
\centering
\caption{Summary of hash-grid-based NeRFs}
\label{tab:hash_grid_nerf_models}
\begin{tblr}{
colspec = {Q[100,c]Q[40,c]Q[20,c]Q[350,l]Q[130,l]},
}
\hline
\textbf{Model} & \textbf{Sensor} & \textbf{Year} & \textbf{Description} & \textbf{Datasets} \\ \hline
MINE \cite{mine2021} & Camera & 2021 & Performs dense 3D reconstruction and novel view synthesis from a single image using a hash-grid encoding. & KITTI \cite{Geiger2013IJRR} \\ \hline
SUDS \cite{turki2023suds} & Camera & 2023 & Extends NeRFs to dynamic large-scale urban scenes, factorising scenes into separate hash table data structures for efficient encoding. & KITTI \cite{Geiger2013IJRR}, V-KITTI 2 \cite{cabon2020vkitti2} \\ \hline
StreetSurf \cite{guo2023streetsurf} & Camera & 2023 & Multi-view implicit surface reconstruction technique tailored for street view images, dividing the scene into close-range, distant-view, and sky regions. & Waymo \cite{Sun_2020_CVPR} \\ \hline
MARS \cite{wu2023mars} & Camera & 2023 & Instance-aware, modular, and realistic simulator for autonomous driving, leveraging hash-grid encoding to model foreground instances and background environments separately. & KITTI \cite{Geiger2013IJRR}, V-KITTI 2 \cite{cabon2020vkitti2} \\ \hline
EmerNeRF \cite{yang2023emernerf} & Camera & 2023 & Self-supervised approach for learning spatial-temporal representations of dynamic driving scenes, capturing scene geometry, appearance, motion, and semantics. & NOTR \cite{yang2023emernerf} \\ \hline
DGNR \cite{li2023dgnr} & Camera & 2023 & Density-Guided Neural Rendering learns a density space to guide the construction of a point-based renderer, eliminating the need for geometric priors. & KITTI \cite{Geiger2013IJRR} \\ \hline
LightSim \cite{pun2023neural} & Camera & 2023 & Neural lighting simulation system for urban driving scenes, generating diverse, controllable, and realistic camera data with accurate illumination and shadows. & PandaSet \cite{10.1109/ITSC48978.2021.9565009} \\ \hline
NeRF-LiDAR \cite{zhang2023nerf} & Lidar & 2023 & Leverages real-world information to generate realistic Lidar point clouds by reconstructing 3D scenes using multi-view images and sparse Lidar data. & nuScenes \cite{nuscenes} \\ \hline
LiDAR-NeRF \cite{tao2023lidar} & Lidar & 2023 & Differentiable end-to-end Lidar rendering framework that synthesises novel views for Lidar sensors, learning geometry and attributes of 3D points. & KITTI-360 \cite{Liao2022PAMI}, NeRF-MVL \cite{tao2023lidar} \\ \hline
NFL \cite{Huang2023nfl} & Lidar & 2023 & Optimises a neural field scene representation from Lidar measurements to synthesise realistic Lidar scans from novel viewpoints. & Waymo \cite{Sun_2020_CVPR} \\ \hline
DyNFL \cite{zheng2024lidar4d} & Lidar & 2023 & Neural field-based approach for high-fidelity re-simulation of Lidar scans in dynamic driving scenes, constructing an editable neural field. & Waymo \cite{Sun_2020_CVPR}, KITTI \cite{Geiger2013IJRR} \\ \hline
UniSim \cite{yang2023unisim} & Camera \& Lidar & 2023 & Closed-loop neural sensor simulation system for self-driving, utilising hash-grid encoding to build neural feature grids for reconstructing scenes. & PandaSet \cite{10.1109/ITSC48978.2021.9565009} \\ \hline
NeuRAD \cite{neurad} & Camera \& Lidar & 2023 & Neural rendering method designed for dynamic automotive scenes, with extensive sensor modelling for both cameras and Lidar. & nuScenes \cite{nuscenes}, PandaSet \cite{10.1109/ITSC48978.2021.9565009}, Argoverse 2 \cite{NEURIPS_DATASETS_AND_BENCHMARKS2021_4734ba6f}, KITTI \cite{Geiger2013IJRR}, ZOD \cite{zod2023} \\ \hline
AlignMiF \cite{tang2024alignmif} & Camera \& Lidar & 2024 & Geometry-aligned multimodal implicit field for joint Lidar-camera synthesis, implementing Geometry-Aware Alignment (GAA) and Shared Geometry Initialisation (SGI). & KITTI-360 \cite{Liao2022PAMI}, Waymo \cite{Sun_2020_CVPR} \\ \hline
\end{tblr}
\end{table*}
\subsection{Hash-Grids}
Hash-grid encoding \cite{mueller2022instant} is the largest subcategory of NeRFs as it is the most effective and computationally efficient method for scene representation. This encoding technique utilises a multi-resolution hash table to map input coordinates to feature vectors efficiently. In hash-grid encoding, a spatial point $\mathbf{p} \in \mathbb{R}^3$ is transformed by hashing its coordinates at multiple resolutions, enabling the network to capture both fine and coarse details. The transformation is defined as follows:
\begin{equation}
    \mathbf{h}_i(\mathbf{p}) = \text{hash}(\mathbf{p} \cdot \mathbf{R}_i) \mod T,
\end{equation}
where $\mathbf{h}_i$ is the hashed index for the $i^{th}$ resolution, $\mathbf{R}_i \in \mathbb{R}^3$ is a resolution-specific scaling factor, and $T \in \mathbb{N}$ is the size of the hash table. Each hashed index $\mathbf{h}_i$ points to a feature vector in the hash table, which is then used to interpolate the final feature representation for the input point $\mathbf{p}$. The summary of hash-grid-based NeRFs is provided in Table \ref{tab:hash_grid_nerf_models}. In the following, we review these models for camera, Lidar, and joint camera and Lidar simulation.

\subsubsection{RGB Image Synthesis}
The pioneering work, MINE \cite{mine2021},  introduces a continuous depth generalisation of the MultiPlane Images (MPI) approach by integrating NeRFs. This method performs dense 3D reconstruction and novel view synthesis from a single image using a hash-grid encoding. By predicting RGB and volume density values at arbitrary depth planes, MINE effectively reconstructs the 3D scene within the camera frustum, filling in occluded contents.
\par
SUDS \cite{turki2023suds} extends NeRFs to dynamic large-scale urban scenes, addressing the limitations of prior methods that typically reconstruct single video clips of short durations. SUDS introduces a scalable representation by factorising scenes into three separate hash table data structures to efficiently encode static, dynamic, and far-field radiance fields. This approach uses unlabelled target signals, including RGB images, sparse Lidar, self-supervised 2D descriptors, and optical flow, enabling photometric, geometric, and feature-metric reconstruction losses. SUDS decomposes dynamic scenes into static backgrounds, individual objects, and their motions, scaling up to tens of thousands of objects across 1.2 million frames from 1700 videos. 
\par
StreetSurf \cite{guo2023streetsurf} presents a multi-view implicit surface reconstruction technique tailored for street view images. Using hash-grid encoding and geometric priors from monocular models, this method addresses the challenges posed by unbounded street views captured with non-object-centric, long, and narrow camera trajectories. The key contribution includes dividing the scene into close-range, distant-view, and sky regions with aligned cuboid boundaries, facilitating fine-grained and disentangled representation. 

\par

\begin{figure}[!t]
\centerline{\includegraphics[width=\linewidth]{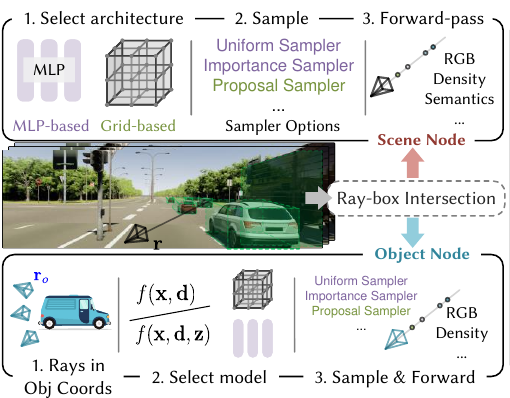}}
\caption{Decomposed ray intersection pipeline of MARS \cite{wu2023mars}, a hash-grid-based NeRF for RGB image synthesis. }
\label{fig:hash_nerf}
\end{figure}

MARS \cite{wu2023mars} introduces an instance-aware, modular, and realistic simulator for ADS, leveraging hash-grid encoding to model foreground instances and background environments separately, as shown in Fig. \ref{fig:hash_nerf}. This modular framework supports flexible switching between different NeRF-related backbones, sampling strategies, and input modalities. MARS achieves state-of-the-art photo-realism by decomposing scenes into foreground and background components, allowing for independent control over static and dynamic properties of instances.
\par
EmerNeRF \cite{yang2023emernerf} introduces a self-supervised approach for learning spatial-temporal representations of dynamic driving scenes. This method stratifies scenes into static and dynamic fields and further parameterises an induced flow field from the dynamic field to aggregate multi-frame features, enhancing the rendering precision of dynamic objects. EmerNeRF's key contribution is its ability to capture scene geometry, appearance, motion, and semantics without relying on ground-truth annotations or pre-trained models. 
\par
DGNR \cite{li2023dgnr} introduces Density-Guided Neural Rendering, which learns a density space from scenes to guide the construction of a point-based renderer. This method eliminates the need for geometric priors by intrinsically learning them from the density space through volumetric rendering. DGNR employs a differentiable renderer to synthesise images from neural density features and uses a density-based fusion module along with geometric regularisation to optimise the density space.
\begin{table*}[!t]
\centering
\caption{Summary of point-based and multi-planar NeRFs}
\label{tab:point_based_multipanar_nerf_models}
\begin{tblr}{
colspec = {Q[80,c]Q[30,c]Q[20,c]Q[350,l]Q[80,l]},
}
\hline
\textbf{Model} & \textbf{Sensor} & \textbf{Year} & \textbf{Description} & \textbf{Datasets} \\ \hline
NPLF \cite{ost2022pointlightfields} & Camera & 2021 & Encodes light fields on sparse point clouds, enabling efficient novel view synthesis for large-scale driving scenarios. & Waymo \cite{Sun_2020_CVPR} \\ \hline
Chang et al. \cite{10378417} & Camera & 2023 & Integrates Lidar maps and 2D conditional GANs to improve novel view synthesis in outdoor environments. & Argoverse 2 \cite{NEURIPS_DATASETS_AND_BENCHMARKS2021_4734ba6f} \\ \hline
DNMP \cite{lu2023dnmp} & Camera & 2023 & Mesh-based rendering combined with neural representations for efficient urban-level radiance field construction. & KITTI-360 \cite{Liao2022PAMI}, Waymo \cite{Sun_2020_CVPR} \\ \hline
LiDAR4D \cite{zheng2024lidar4d} & Lidar & 2024 & Differentiable Lidar-only framework for novel space-time Lidar view synthesis, using a 4D hybrid representation. & KITTI-360 \cite{Liao2022PAMI}, NuScenes \cite{nuscenes} \\ \hline
\end{tblr}
\end{table*}
\par
LightSim \cite{pun2023neural} presents a neural lighting simulation system for urban driving scenes, enabling the generation of diverse, controllable, and realistic camera data. LightSim reconstructs lighting-aware digital twins from real-world sensor data, including geometry, appearance, and estimated scene lighting. This system facilitates actor insertion, modification, removal, and relighting from new viewpoints with accurate illumination and shadows.
\subsubsection{Lidar Point Cloud Synthesis}
NeRF-LiDAR \cite{zhang2023nerf} presents a novel Lidar simulation method that leverages real-world information to generate realistic Lidar point clouds. This method reconstructs 3D scenes using multi-view images and sparse Lidar data collected by self-driving cars. By learning the NeRF representation for real-world scenes, NeRF-LiDAR generates point clouds with accurate semantic labels, which significantly boosts the performance of 3D segmentation models trained on this simulated data. 
\par
LiDAR-NeRF \cite{tao2023lidar} introduces the first differentiable end-to-end Lidar rendering framework, designed to synthesise novel views for Lidar sensors. This method leverages neural radiance fields to jointly learn geometry and attributes of 3D points, such as intensity and ray-drop probability, without explicit 3D reconstruction. The approach incorporates structural regularisation to preserve local details, significantly improving the accuracy of synthesised Lidar patterns.
\par
NFL model \cite{Huang2023nfl} optimises a neural field scene representation from Lidar measurements to synthesise realistic Lidar scans from novel viewpoints. NFL combines the rendering power of neural fields with a detailed, physically motivated model of the Lidar sensing process, accurately reproducing key sensor behaviours like beam divergence, secondary returns, and ray dropping. The key innovation lies in integrating these physical characteristics into the neural field framework, leading to improved realism in synthesising Lidar views.
\par
DyNFL model \cite{zheng2024lidar4d} introduces a novel neural field-based approach for high-fidelity re-simulation of Lidar scans in dynamic driving scenes. This method processes Lidar measurements from dynamic environments, constructing an editable neural field by separately reconstructing static backgrounds and dynamic objects. DyNFL utilises a neural field composition technique that integrates reconstructed neural assets from various scenes through a ray drop test, accounting for occlusions and transparent surfaces. This approach significantly improves the physical fidelity and flexible editing capabilities of dynamic scene Lidar simulation.
\par
\begin{figure*}[!t]
\centerline{\includegraphics[width=\linewidth]{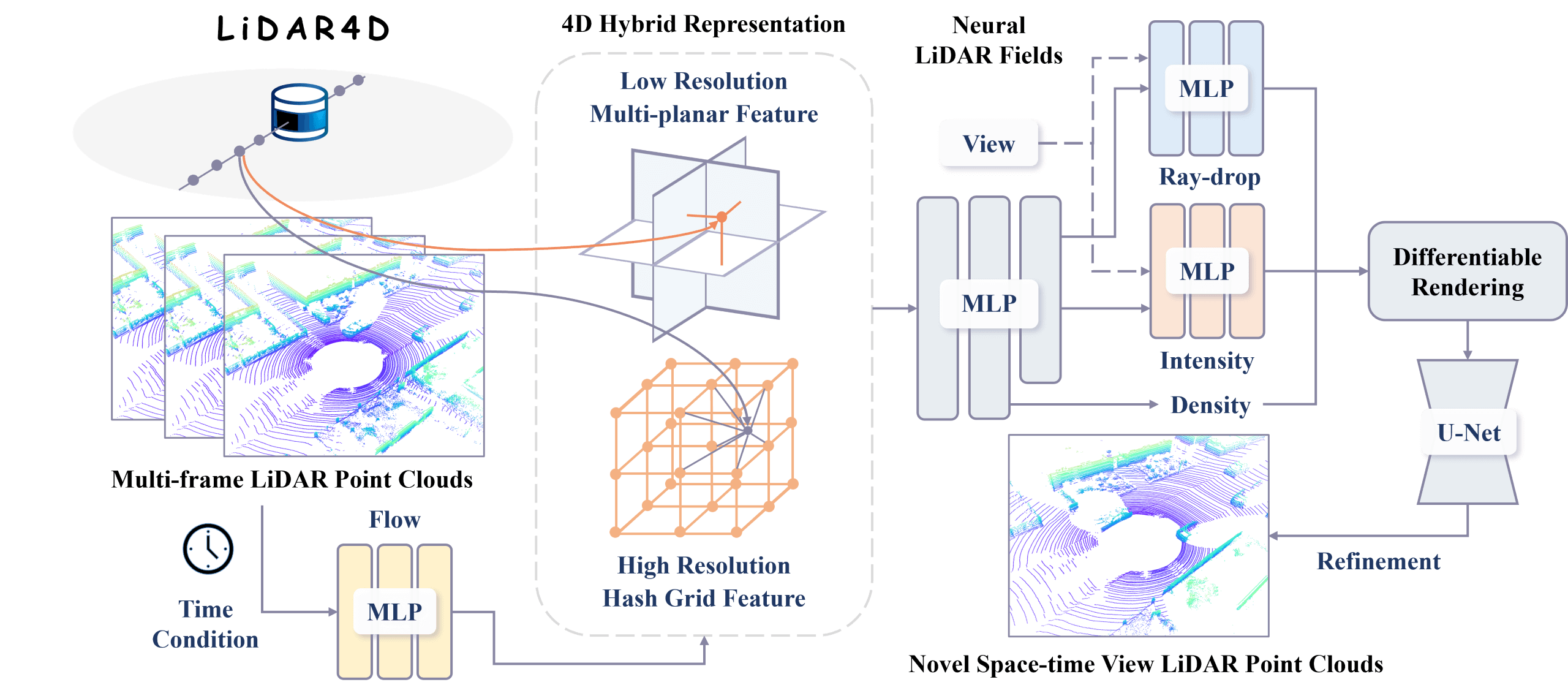}}
\caption{Overview of LiDAR4D \cite{zheng2024lidar4d} model, a NeRF model which uses both multi-planar and hash-grid features for Lidar point cloud simulation. }
\label{fig:palanar_nerf}
\end{figure*}

\subsubsection{Joint Camera and Lidar Simulation}
UniSim \cite{yang2023unisim} presents a closed-loop neural sensor simulation system for self-driving, utilising hash-grid encoding to build neural feature grids that reconstruct both static backgrounds and dynamic actors in the scene. This method enables the generation of realistic multi-sensor simulations from a single recorded log, allowing for scene manipulation and the creation of new scenarios. UniSim integrates learnable priors for dynamic objects and a convolutional network to handle unseen regions, improving the realism of extrapolated views.
\par
NeuRAD \cite{neurad} is a neural rendering method specifically designed for dynamic automotive scenes. It features a simple network design and extensive sensor modelling for both cameras and Lidar, including effects such as rolling shutter, beam divergence, and ray dropping. NeuRAD employs a unified neural feature field for both static and dynamic elements, enabling realistic sensor data generation and the editing of the pose of the ego vehicle and other road users.
\par
AlignMiF \cite{tang2024alignmif} introduces a geometry-aligned multimodal implicit field for joint Lidar-camera synthesis, addressing the challenge of misalignment between different sensor modalities. By implementing Geometry-Aware Alignment (GAA) and Shared Geometry Initialisation (SGI), AlignMiF aligns the coarse geometry across Lidar and camera data, significantly enhancing the fusion process.

\subsection{Point-based and Multi-planar Encoding}
Point-based and multi-planar NeRFs are advanced techniques designed to improve the efficiency and quality of scene representation and reconstruction in NeRFs. In point-based NeRFs, the scene is represented using a sparse set of 3D points, often captured by Lidar sensors. Each point is associated with feature descriptors that are used to interpolate light fields for rendering. This approach reduces the computational burden associated with traditional volumetric models and enables efficient processing of large-scale environments. Multi-planar encoding extends this concept by utilising multiple planes to represent the scene at different depths. The summary of point-based and multi-planar NeRFs is provided in Table \ref{tab:point_based_multipanar_nerf_models}. In the following, we review these models for camera and Lidar simulation.
\par
\begin{table*}[!t]
\centering
\caption{Summary of 3D Gaussian splatting models (all used for camera RGB image synthesis).}
\label{tab:3dgs_models}
\begin{tblr}{
colspec = {Q[100,c]Q[20,c]Q[400,l]Q[70,l]},
}
\hline
\textbf{Model} & \textbf{Year} & \textbf{Description} & \textbf{Datasets} \\ \hline
PVG \cite{chen2023periodic} & 2023 & Unified representation model capturing static and dynamic elements using periodic vibration-based temporal dynamics and adaptive control for efficient large scene representation. & Waymo \cite{Sun_2020_CVPR}, KITTI \cite{Geiger2013IJRR} \\ \hline
Street Gaussians \cite{yan2023streetgaussians} & 2023 & Explicit scene representation modelling dynamic urban streets from monocular videos, separating foreground vehicles from the static background, enabling real-time rendering. & KITTI, Waymo \cite{Sun_2020_CVPR} \\ \hline
DrivingGaussian \cite{zhou2024drivinggaussian} & 2024 & Efficient framework for reconstructing and rendering dynamic scenes, using incremental static 3D Gaussians and a composite dynamic Gaussian graph for multiple moving objects. & nuScenes \cite{nuscenes}, KITTI-360 \cite{Liao2022PAMI} \\ \hline
HUGS \cite{zhou2024hugs} & 2024 & Comprehensive framework for urban scene understanding via 3D Gaussian splatting, optimising geometry, appearance, semantics, and motion for real-time novel view synthesis. & KITTI \cite{Geiger2013IJRR}, KITTI-360 \cite{Liao2022PAMI}, V-KITTI 2 \cite{cabon2020vkitti2} \\ \hline
SGD \cite{yu2024sgd} & 2024 & Enhances street view synthesis by leveraging a diffusion model for prior knowledge and 3D Gaussian splatting to improve rendering quality from sparse training views. & KITTI \cite{Geiger2013IJRR}, KITTI-360 \cite{Liao2022PAMI}\\ \hline
DeSiRe-GS~\cite{peng2024desiregs4dstreetgaussians} & 2024 & Self-supervised 4D Gaussian splatting model decomposing static and dynamic elements with temporal consistency and motion masks for accurate scene reconstruction.& KITTI \cite{Geiger2013IJRR}, Waymo \cite{Sun_2020_CVPR}\\ \hline
\end{tblr}
\end{table*}
NPLF \cite{ost2022pointlightfields} introduces a novel representation that encodes light fields on sparse point clouds, enabling efficient novel view synthesis for large-scale driving scenarios. This method departs from traditional volumetric models by leveraging the geometric properties of point clouds, which are sparsely captured by Lidar sensors. NPLF computes per-point features using a learned feature extractor and interpolates these features to form light field descriptors for each ray, ensuring consistent and unique descriptions. This technique significantly reduces the computational burden of volumetric sampling and achieves realistic novel view synthesis with minimal training data. 
\par
Chang et al. \cite{10378417} introduce a point-based NeRF framework that integrates Lidar maps and 2D conditional GANs (cGANs) to improve novel view synthesis in outdoor environments. This method leverages the strong 3D geometry priors provided by Lidar sensors, significantly enhancing ray sampling locality and reducing artefacts caused by imperfect Lidar data. By using Lidar maps as sparse samples of the environment, the system assigns localised embeddings and performs volume rendering. The integration of cGANs refines the synthesised images, producing higher quality outputs.
\par

Lu et al. \cite{lu2023dnmp} propose a new framework, DNMPs, for efficient urban-level radiance field construction. DNMPs leverage mesh-based rendering combined with neural representations to model both geometry and radiance information compactly. Each DNMP consists of connected deformable mesh vertices paired with radiance features, optimising shape and radiance using a latent code. This approach significantly reduces computational costs and memory usage while maintaining high-quality rendering.
\begin{figure*}[!t]
\centerline{\includegraphics[width=\linewidth]{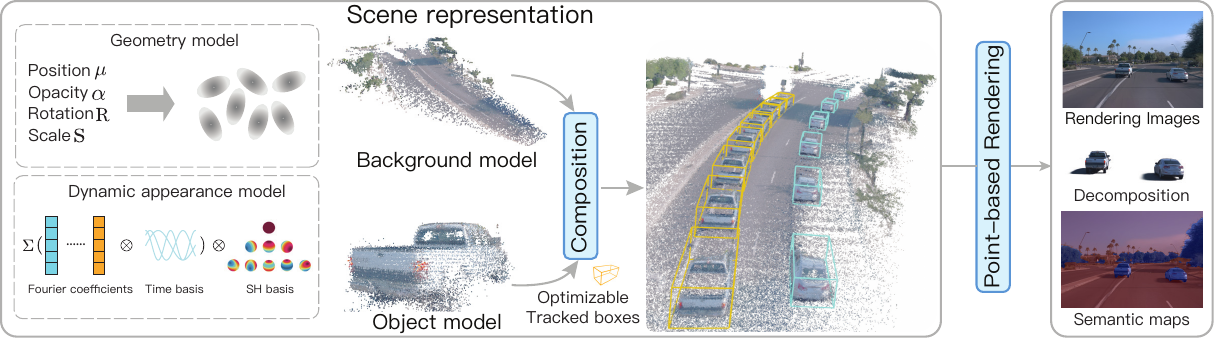}}
\caption{Overview of rendering pipeline in Street Guassians \cite{yan2023streetgaussians}, a volume renderer based on 3D Gaussian Splatting for RGB image synthesis. }
\label{fig:3dgs}
\end{figure*}
\par
LiDAR4D \cite{zheng2024lidar4d} introduces a differentiable Lidar-only framework for novel space-time Lidar view synthesis. This method tackles the challenges of dynamic scene reconstruction by using a 4D hybrid representation that combines multi-planar and grid features, as shown in Fig. \ref{fig:palanar_nerf}. LiDAR4D incorporates geometric constraints derived from point clouds to improve temporal consistency and uses global optimisation of ray-drop probability to preserve cross-region patterns.

\section{Volume Renderers -- 3D Gaussian Splatting} \label{sec:3DGS}
3D Gaussian Splatting (3DGS) models have emerged as a powerful technique for scene rendering in ADS. These models utilise 3D Gaussian functions to efficiently capture and represent both static and dynamic elements in urban environments. By incorporating mechanisms that enhance temporal continuity and leverage adaptive control strategies, 3DGS models ensure detailed and consistent scene representations. They employ point-based encoding of the scene, using incremental and composite approaches to model static backgrounds and dynamic objects. Compared to NeRFs, 3DGS offers significant advantages in rendering speed, achieving up to 900-fold acceleration \cite{chen2023periodic}, which is crucial for real-time ADS applications. It also demonstrates superior performance in reconstruction quality and novel view synthesis, particularly in dynamic urban scenes, as evidenced by improved PSNR, SSIM, and LPIPS metrics. However, 3DGS faces challenges in maintaining temporal consistency in dynamic scenes and may struggle with fine geometric details, areas where NeRFs might perform better due to their continuous representation. These trade-offs between rendering speed, image quality, and geometric accuracy are important considerations when applying 3DGS models in ADS scenarios. The summary of 3DGS models is provided in Table \ref{tab:3dgs_models}. In the following, we review these models that are all used to synthesise RGB images.
\par
PVG method \cite{chen2023periodic} introduces a unified representation model designed to efficiently and uniformly capture both static and dynamic elements in large-scale urban scenes. Building on the 3D Gaussian splatting technique, PVG incorporates periodic vibration-based temporal dynamics, enabling a cohesive representation of scene characteristics. The model introduces a temporal smoothing mechanism and a position-aware adaptive control strategy to enhance temporal continuity and large scene representation learning with sparse training data.
\par
DrivingGaussian \cite{zhou2024drivinggaussian} presents an efficient framework for reconstructing and rendering surrounding dynamic scenes in autonomous driving contexts. This method sequentially models static backgrounds with incremental static 3D Gaussians and uses a composite dynamic Gaussian graph to handle multiple moving objects. By leveraging Lidar priors for Gaussian splatting, DrivingGaussian reconstructs scenes with high detail and maintains panoramic consistency across multi-camera setups. The model outperforms existing methods in dynamic driving scene reconstruction, achieving photo-realistic surround-view synthesis with high fidelity and consistency. 
\par
Street Gaussians \cite{yan2023streetgaussians} introduces an explicit scene representation designed to efficiently model dynamic urban street scenes from monocular videos. This method represents dynamic urban streets as point clouds equipped with semantic logits and 3D Gaussians, separating foreground vehicles from the static background (see Fig. \ref{fig:3dgs}). By optimising tracked poses and utilising a dynamic spherical harmonics model for the dynamic appearance of vehicles, Street Gaussians can efficiently compose object vehicles and backgrounds.
\par
HUGS \cite{zhou2024hugs} proposes a comprehensive framework for urban scene understanding from RGB images. By utilising 3DGS, HUGS jointly optimises geometry, appearance, semantics, and motion, facilitating real-time novel view synthesis and dynamic scene reconstruction. This method incorporates physical constraints for moving object poses, improving accuracy even with noisy 3D bounding box predictions.
\par
Yu et al. \cite{yu2024sgd} present a novel approach for enhancing street view synthesis in ADS simulations. This method leverages a DM to provide prior knowledge, combined with 3DGS to improve rendering quality from sparse training views. By fine-tuning the DM with images from adjacent frames and depth data from Lidar point clouds, the model regularises 3DGS training and enhances the quality of novel view synthesis. 
\par
DeSiRe-GS~\cite{peng2024desiregs4dstreetgaussians} extends 3D Gaussian splatting to a 4D representation for self-supervised static-dynamic decomposition and surface reconstruction in urban driving scenarios. By leveraging 2D motion masks and temporal consistency, it models dynamic elements as time-varying variables and reconstructs Gaussian surfaces without explicit 3D annotations. The method achieves state-of-the-art scene decomposition and reconstruction performance, effectively addressing data sparsity challenges.

\begin{figure}[!t]
\centerline{\includegraphics[width=\columnwidth]{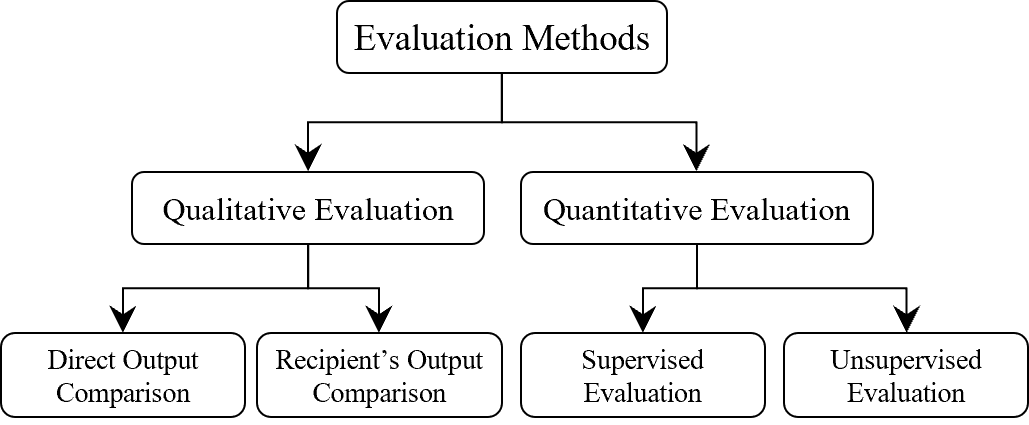}}
\caption{Categorising validation approaches of perception sensor simulation models.}
\label{fig:cat_evaluation}
\end{figure}
\begin{figure*}[!t]
\centerline{\includegraphics[width=\linewidth]{figure/qualitative-comparison.pdf}}
\caption{Qualitative evaluation of a Lidar simulation model by direct comparison. The synthesised point cloud of LiDAR-NeRF \cite{tao2023lidar} is compared to state-of-the-art models and ground-truth.}
\label{fig:eval_qualitative}
\end{figure*}
\section{Evaluation Approaches} \label{sec:eval}  \label{sec:sensor_model_validation}
Similar to any other modelling approach, it is crucial to thoroughly evaluate the performance of simulation models concerning their intended role within the system \cite{OBERKAMPF2008716}.  In the evaluation of perception sensor simulation models for ADS, two general methodologies, namely qualitative and quantitative evaluation, are commonly employed. Qualitative evaluation involves visualising either the output of the model itself or the output of a recipient component, such as a perception model. On the other hand, quantitative evaluation entails assessing the model based on either supervised metrics, when reference data is available, or unsupervised metrics, when reference data is not accessible. The categorisation of the evaluation approaches is depicted in Fig. \ref{fig:cat_evaluation}.
In the subsections below, we will explore existing examples of both qualitative and quantitative evaluations in the literature of camera and Lidar simulation models for ADS. 

\subsection{Qualitative Evaluation}
Qualitative evaluation is the most common approach for assessing the sensor simulation models. This process involves visualising the output of the sensor model and comparing it to state-of-the-art methods and real data. Moreover, some studies emphasise visualising the output of a perception model, such as a semantic segmentation network. A more detailed exploration of these two qualitative evaluation approaches is provided in the subsequent paragraphs.
\subsubsection{Direct Output Visualisation}
Most camera and Lidar models visualise their synthesised outputs and compare them with real data and state-of-the-art methods. The input data for sensor simulation models is also visualised (except in unconditional approaches) to understand the information the model learns from. For example, in the sim-to-real mapping network introduced by \cite{9756256}, various simulated buffers such as normal, depth, and albedo, are processed and then visualised during the evaluation phase. For comparison with real sensory data, the model's output is either compared to ground-truth data in cases of supervised methods (see Fig. \ref{fig:eval_qualitative}), or with nearest neighbour samples from the training dataset regarding unsupervised approaches.  In examples such as the LidarSim \cite{manivasagam2020lidarsim} framework, the entire scene is reconstructed using real data, providing corresponding ground-truth for each Lidar viewpoint. Conversely, Dusty \cite{Nakashima2021LearningTD} generates data randomly, and the nearest neighbour samples from real data are used for comparison. Regarding data representation, camera models often visualise synthesised RGB images, whereas Lidar models visualise BEV images, range images or 3D point clouds based on the utilised data representation. \par
Additionally, learned models are sometimes applied to applications such as data restoration, and the results are visualised to demonstrate the model’s effectiveness.
For example, Dusty2 \cite{nakashima2022generative} model was used to reconstruct real range images or restore distorted images. The high-quality restored or reconstructed images demonstrate the model's ability to cover data distribution effectively.
\begin{figure}[!t]
\centerline{\includegraphics[width=\columnwidth]{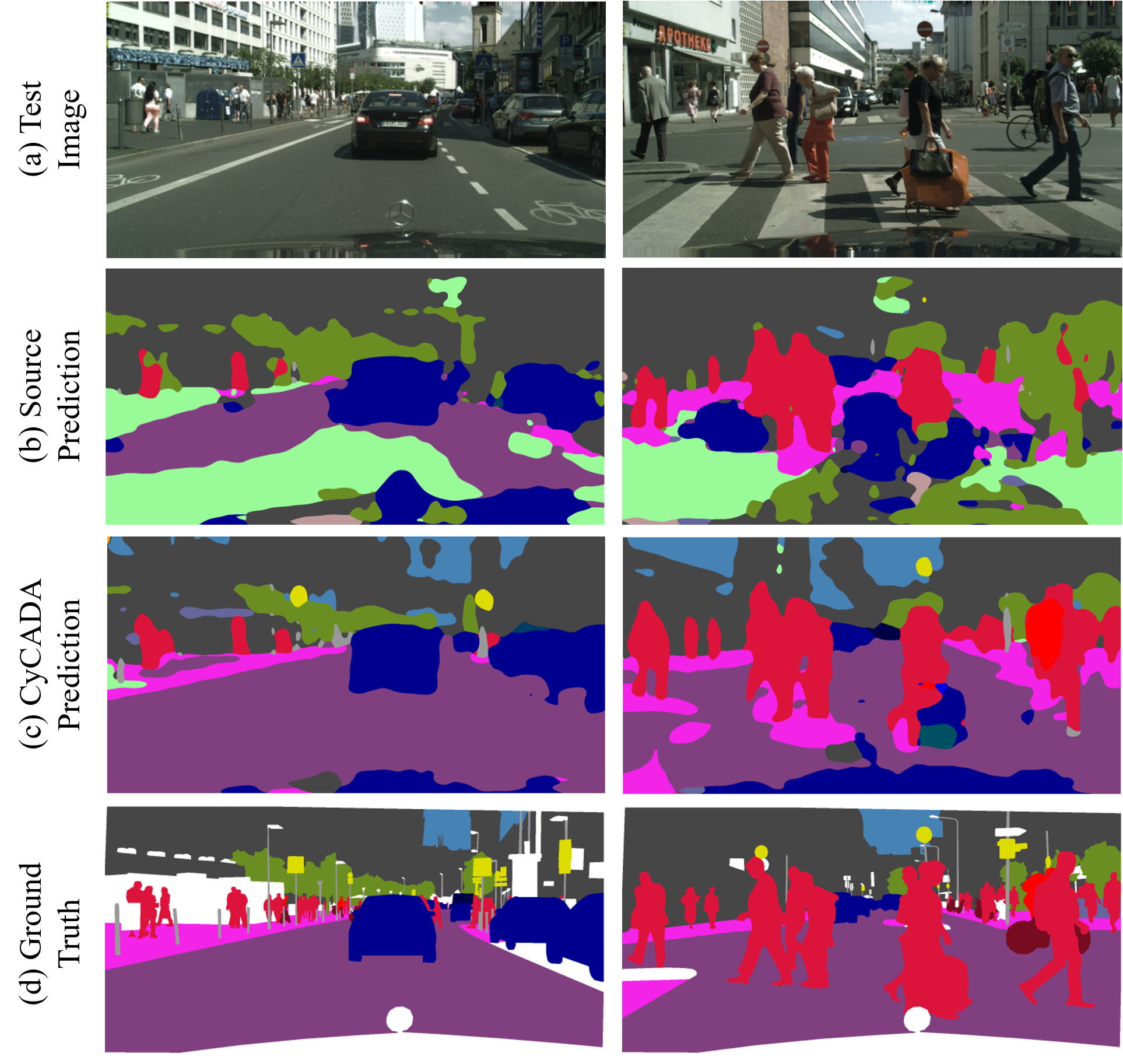}}
\caption{Qualitative validation of CyCADA \cite{pmlr-v80-hoffman18a} by comparing the output of a downstream task, e.g. semantic segmentation network. The semantic segmentation network is trained with two different image sources: (1) synthetic images of the GTA-V dataset, and (2) the modelled ones by CyCADA. The test images (a) are fed to these networks and the prediction is depicted in (b) and (c), respectively. The last row (d) shows the ground-truth layout.}
\label{fig:val_recipient}
\end{figure}
\subsubsection{Recipient's Output Comparison}
As perception sensors are not standalone systems in ADS, it is essential to consider their integration with downstream components that receive sensory data. Considering a data-driven model as the recipient component, several studies \cite{pmlr-v80-hoffman18a, 9339933, Guillard2022, manivasagam2020lidarsim} have visualised the predicted annotations of the models. This approach allows for evaluating the performance of the models that have been trained on synthesised data using real test data. The closer the output of the sensor model resembles the real-world data, the higher the accuracy of the perception model in detecting and interpreting its input. Fig.~\ref{fig:val_recipient} demonstrates an example of CyCADA's \cite{pmlr-v80-hoffman18a} evaluation through predictions of a semantic segmentation network, FCN \cite{7478072}.  As shown, the FCN model is much more accurate when trained on CyCADA's synthesised images than on the source images. This demonstrates the effectiveness of CyCADA in improving the realism of GTA-V simulated images.


\subsection{Quantitative Evaluation}
Quantitative evaluation of the sensor simulation models entails establishing a mathematical metric to measure the model's performance against the expected output. We categorise these quantitative evaluation methods into supervised and unsupervised groups. Supervised evaluation techniques require paired model input and ground-truth output for the assessment, while unsupervised evaluation techniques do not enforce such necessity. We will explain the quantitative evaluation techniques in more detail in the following paragraphs.
\subsubsection{Supervised Evaluation}
To quantitatively evaluate a sensor model in a supervised manner, it is necessary to establish correspondence between the synthesised sensory data and the real-world data. The supervised evaluation methods are often utilised in supervised modelling approaches where the models are trained with a pair of input-output data, thus enabling comparison with ground-truth. For instance,
in the cases where semantic summation layout is transformed into RGB images \cite{Isola2016ImagetoImageTW} or Lidar intensity is predicted using the spatial coordinates \cite{9339933}, the ground-truth RGB or intensity image in the test set is used for the supervised evaluation. Similarly, other frameworks such as SurfelGAN \cite{Yang_2020_CVPR} and LidarSim \cite{manivasagam2020lidarsim} reconstruct the 3D scene using real sensory data frames, thus providing access to the expected models' output at specific viewpoints of the sensor. \par
Supervised evaluation metrics commonly focus on calculating the average per-pixel error in image-based sensory data representations. These metrics include  Root Mean Squared Error (RMSE, i.e. L2 distance error), Mean Absolute Error (MAE, L1 distance error), thresholded accuracy (ratio of pixels with the error less than a certain threshold), and PSNR. For calculating the distance of the Lidar point cloud in 3D representation, Chamfer Distance (CD) and Earth Mover's Distance (EMD) \cite{Fan2016APS} are frequently utilised. Additionally, other supervised metrics such as SSIM and LPIPS \cite{Zhang2018TheUE} are used, relying on structural and feature similarity rather than solely on pixel-level error.

\subsubsection{Unsupervised Evaluation}
Building an environment model that accurately reflects complex real-world scenarios is a challenging task. Therefore, there is often no direct correspondence between real and synthesised sensory data, thus unsupervised evaluation methods are commonly employed. These methods typically rely on constructing two sets of synthesised and real data, considering them as distributions, and employing metrics to calculate the distance between these distributions. Frechet Inception Distance (FID) \cite{Heusel2017GANsTB}, Kernel Inception Distance (KID) \cite{bińkowski2018demystifying}, and Sliced Wasserstein Distance (SWD) \cite{Karras2017ProgressiveGO} are common metrics used to measure this distance for the sensory data represented as images. In the work by Ritcher et al. \cite{9756256}, the authors proposed semantically aligned Kernel VGG distance (sKVD) that addresses the bias toward semantic similarity of the scenes, a limitation often found in FID/KID. Concerning finding the distance between distributions of two Lidar 3D point clouds, Jensen-Shannon Divergence (JSD) and Minimum Matching distance (MMD) have also been employed \cite{Nakashima2021LearningTD}. In assessing the performance of unconditional modelling approaches, metrics such as Coverage (COV) and 1-Nearest Neighbour Accuracy (1-NNA) have also been used \cite{Nakashima2021LearningTD} to measure the diversity of generated samples. \par
\begin{table}
\centering
\caption{Summary of quantitative evaluation metrics.}
\label{tab:eval_metrics}
\begin{tblr}{colspec={ l l X[l, 3cm] },
  colsep = 1.7pt,
  cell{2}{1} = {r=3}{c},
  cell{5}{1} = {r=3}{c},
  hline{1-2,5,9} = {-}{},
}
\textbf{Category} & \textbf{Description}          & \textbf{Metric}            \\
Supervised                 & pixel-level error measurement       & RMSE, MAE, PSNR            \\
                           & perceptual similarity               & SSIM, LPIPS \cite{Zhang2018TheUE}                \\
                           & 3D point cloud distance             & CD, EMD \cite{Fan2016APS}                    \\
Unsupervised               & distribution distance (images)       & FID \cite{Heusel2017GANsTB}, KID \cite{bińkowski2018demystifying}, sKVD \cite{9756256}, SWD  \cite{Karras2017ProgressiveGO}       \\
                           & distribution distance (point clouds) & JSD \cite{Fan2016APS}, MMD \cite{Fan2016APS}                  \\
                           & distribution diversity & COV \cite{Fan2016APS}, 1-NNA \cite{Fan2016APS}                  \\
                           & perception model's performance      & IOU, mAP, pixAcc, classAcc 
\end{tblr}
\end{table}
Some studies have also incorporated human perceptual studies \cite{Chen2017PhotographicIS} to measure the realism of the synthesised RGB images. This evaluation involves visualising the paired synthesised and real images to the human users and reporting the percentage of times when users preferred the synthesised image. The studies commonly used the Amazon Mechanical Turk (AMT) platform for conducting evaluations.\par 
 The most widely used unsupervised evaluation approach found in the literature involves assessing the performance of a downstream perception model. This process starts by conducting a set of synthesised sensory data to train the perception model. After training with the synthesised dataset, the perception model is then validated on a real sensory dataset. The performance of the perception model is compared to its performance when initially trained on real data. The principle of this approach is that the closer the perception model's performance to its performance on real data, the more realistic the synthesised data is perceived by the model. Typically, the perception models used in these evaluations are SOTA object detection or semantic segmentation models. Key performance metrics for these models include Intersection Over Union (IOU), average Precision (AP), average pixel accuracy (pixAcc), and average class accuracy (classAcc). The summary of the quantitative evaluation metrics is provided in Table  \ref{tab:eval_metrics}.

\section{Conclusion and Future Research Directions} \label{sec:conclusion}
This review explored state-of-the-art data-driven camera and Lidar simulation models for ADS development and validation. The paper categorised and reviewed 96 different models from the novel perspective of generative models, including GANs, diffusion models, and auto-regressive models, as well as volume renderers comprising NeRFs and 3DGS. By focusing on these data-driven approaches, the review aimed to provide insights into the rapid expansion of the literature in this field, driven by the growing availability of real-world recorded perception sensory datasets and the success of deep learning models in synthesising high-dimensional data. While these models have shown promising results in enhancing the realism and efficiency of sensor simulation, several challenges and opportunities for future research have emerged.
\par
\textbf{Standardising Evaluation and Benchmarking:}
A critical challenge in current research is the lack of standardised protocols and comprehensive benchmarks for assessing sensor simulation models in ADS applications. To address this limitation, future research should prioritise:
\begin{itemize}
\item Developing robust benchmarks to evaluate simulation models' capability in generating data for edge cases, such as adverse weather conditions, crucial for ADS safety testing.
\item Developing standardised protocols to evaluate the computational efficiency and scalability of simulation models, focusing on their suitability for real-time ADS applications and large-scale data generation.
\item Enhancing evaluation methodologies beyond traditional ego-vehicle trajectory analysis. This expansion should include assessments of novel spatial and temporal view synthesis capabilities, particularly for volume renderers \cite{zheng2024lidar4d}.
\end{itemize}
\par
\textbf{Enhancing Real-time Performance and Controllability:}
The suitability of data-driven models for real-time ADS applications, such as Hardware-in-the-Loop (HiL) testing, remains largely unexplored in the literature. This gap is particularly evident in certain diffusion models \cite{zyrianov2024lidardm}, which currently fall short of real-time capabilities. Furthermore, most of the discussed generative models are either unconditional or rely on simplistic control inputs, which restricts their adequacy for real-world applications, e.g. implementing diverse testing scenarios. These limitations present significant challenges for integrating data-driven models into time-critical ADS testing workflows. To address these issues, future research should focus on:
\begin{itemize}
    \item Optimising deep learning-based models to achieve real-time or faster-than-real-time performance, potentially through hardware-specific implementations and algorithmic innovations.
    \item Accelerating generation algorithms in computationally intensive models, including diffusion models, leveraging techniques such as latent consistency \cite{luo2023latentconsistencymodelssynthesizing}.
    \item Enhancing the controllability of generative models to allow for precise adjustments to synthesised sensory data, enabling researchers to replicate specific testing scenarios, such as varying lighting, weather conditions, and sensor occlusions.
\end{itemize}
\par
\textbf{Improving Trustworthiness and Generalisation:}
Deep learning models inherently struggle to generalise beyond their training data distribution, raising concerns about their reliability in diverse scenarios \cite{Huang2018ASO}. This limitation is particularly pertinent to reviewed data-driven simulation models, which heavily rely on deep learning approaches. Given that these models play a crucial role in ADS testing, their ability to accurately represent a wide range of driving conditions is of paramount importance. To enhance trustworthiness, future research should prioritise:
\begin{itemize}
    \item Exploring hybrid approaches that combine data-driven simulation methods with physics-based models to improve the generalisation capabilities and overall reliability.
    \item Investigating transfer learning approaches to leverage knowledge from high-fidelity and complex models to improve the performance of more efficient, deployable models.
    \item Developing robust methods to quantify and communicate uncertainty in simulated sensor data, enhancing the transparency and reliability of virtual ADS testing environments.
\end{itemize}
\par
\textbf{Enhancing Volume Rendering Techniques:}
Despite significant advancements, volume renderers \cite{neurad,zhou2024hugs,chen2023periodic,yang2023unisim,zheng2024lidar4d} still face several specific challenges in accurately simulating complex driving scenes. To address these limitations, future research should focus on:
\begin{itemize}
    \item Enhancing NeRFs to accurately simulate deformable actors and dynamic objects over extended periods.
    \item Improving 3DGS performance in challenging environmental conditions and night-time scenarios.
    \item Developing techniques for modelling dynamic scene elements and enhancing temporal coherence through scene flow estimation \cite{chen2023periodic}.
    \item Enhancing geometric accuracy for both static and dynamic objects, particularly in handling occlusions and long-distance motions.
\end{itemize}
\par
Data-driven camera and Lidar simulation models have made significant strides, yet critical areas for improvement remain. Addressing these challenges is crucial for enhancing the reliability, efficiency, and fidelity of sensor simulation models in ADS applications. As the field evolves, interdisciplinary collaboration and the development of standardised evaluation methods will be key to unlocking the full potential of these technologies and ensuring their safe, effective deployment in real-world ADS scenarios.

\section*{Acknowledgment}

This research is supported in part by the University of Warwick's Centre for Doctoral Training in Future Mobility Technologies and in part
by the Hi-Drive Project through the European Union’s Horizon 2020 Research and Innovation Program under Grant Agreement No 101006664.
We thank Dr Christoph Kessler from Ford for his valuable comments and suggestions on improving the paper. The sole responsibility of this publication lies with the authors.

\ifCLASSOPTIONcaptionsoff
  \newpage
\fi


\bibliographystyle{IEEEtran} 
\bibliography{references_sensor_model.bib}

%
\vspace{-5pt}

\begin{IEEEbiography}[{\includegraphics[width=1in,height=1.25in,clip,keepaspectratio]{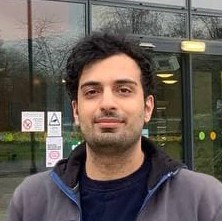}}]{Hamed Haghighi}
is a PhD candidate with the Warwick Manufacturing Group (WMG) at the University
of Warwick, UK. He received a B.Sc. (2016) in Software Engineering from the Isfahan University of Technology (Isfahan, Iran) and an M.Sc. (2019) in Artificial Intelligence from the University of Tehran (Tehran, Iran). His research interests include machine learning, computer
vision, computer graphics, and autonomous vehicles.
\end{IEEEbiography}

\begin{IEEEbiography}[{\includegraphics[width=1in,height=1.25in,clip,keepaspectratio]{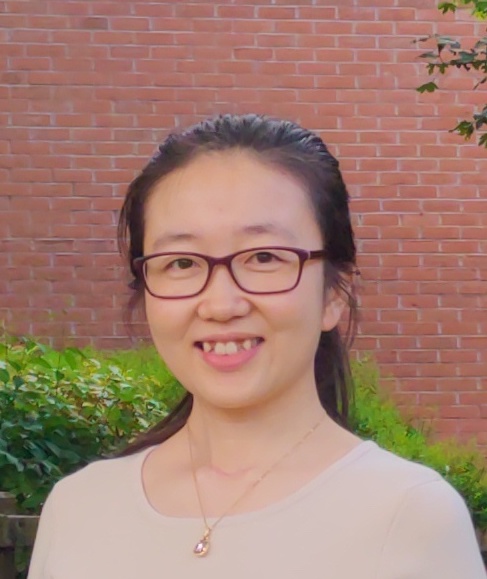}}]{Xiaomeng Wang}
Dr Xiaomeng Wang received a B.S. degree in Communication Engineering and an MSc degree in Information Engineering from Communication University of China in 2010 and 2013 respectively, and a PhD degree in Computer Science from the Computer Vision Laboratory at the University of Nottingham in 2018. Xiaomeng has worked as a research associate at the Graphics \& Interaction Group in the Department of Computer Science and Technology, University of Cambridge, from 2017 to 2019. She is currently a research fellow in the Intelligent Vehicle Group at WMG, University of Warwick. Her main research interests involve computer vision, machine learning, and their applications.  
\end{IEEEbiography}
\begin{IEEEbiography}[{\includegraphics[width=1in,height=1.25in,clip,keepaspectratio]{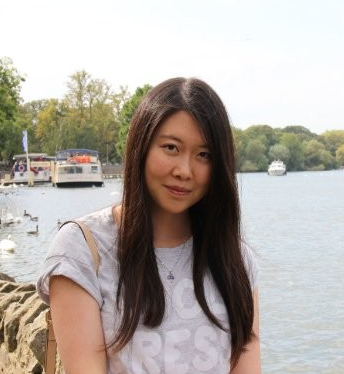}}]{Hao Jing}
Dr Hao Jing is currently a Senior Research Fellow in the Intelligent Vehicle Group at WMG, University of Warwick, with a research focus on high-performance integrated and cooperative vehicle positioning and navigation solutions in challenging environments. Dr Jing previously completed her PhD at the University of Nottingham on the topic of collaborative indoor positioning. She has worked on several projects that focus on achieving reliable and robust navigation performance for Connected and Autonomous Vehicles (CAV), pedestrians and mobile mapping systems in various environments and scenarios, based on solutions that make use of GNSS, Lidar, IMU and wireless signals.
\end{IEEEbiography}

%

\begin{IEEEbiography}[{\includegraphics[width=1in,height=1.25in,clip,keepaspectratio]{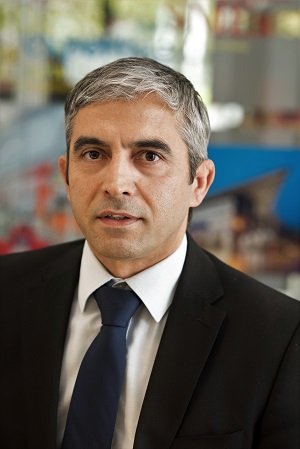}}]{Mehrdad Dianati}
Professor Mehrdad Dianati leads Networked Intelligent Systems (Cooperative Autonomy) research at Warwick Manufacturing Group (WMG), University of Warwick. He has over 28 years of combined industrial and academic experience, with 20 years in leadership roles in multi-disciplinary collaborative R\&D projects, in close collaboration with the Automotive and ICT industries. He is also the Co-Director of Warwick's Centre for Doctoral Training on Future Mobility Technologies, training doctoral researchers in the areas of intelligent and electrified mobility systems in collaboration with the experts in the field of electrification from the Department of Engineering of the University of Warwick. In the past, he served as an associate editor for the IEEE Transactions on Vehicular Technology and several other international journals, including IET Communications. Currently, he is the Field Chief Editor of Frontiers in Future Transportation. His academic experience includes over 25 years of teaching undergraduate and post-graduate level courses and supervision of research students. He currently leads a post-graduate course in Machine Intelligence.
\end{IEEEbiography}




\end{document}